\documentclass[11pt]{article}

\usepackage[preprint]{acl}

\usepackage{times}
\usepackage{latexsym}

\usepackage[T1]{fontenc}

\usepackage[utf8]{inputenc}

\usepackage{microtype}

\usepackage{inconsolata}
\usepackage{booktabs}
\usepackage{multirow}
\usepackage{tcolorbox}
\tcbuselibrary{listingsutf8}
\usepackage{amsmath}
\usepackage{longtable}
\usepackage{listings}

\usepackage{threeparttable}
\usepackage[table,xcdraw]{xcolor} 

\usepackage{graphicx}
\usepackage[colorinlistoftodos]{todonotes}

\title{PExA: Parallel Exploration Agent for Complex Text-to-SQL}

\author{Tanmay Parekh$^1$\thanks{~This work was done during an internship at Bloomberg.} \ \ \ \ \
Ella Hofmann-Coyle$^2$\thanks{~Research performed while working at Bloomberg.} \ \ \ \ \
Shuyi Wang$^2$ \\
{\bf Sachith Sri Ram Kothur$^2$ \ \ \ \ \
Srivas Prasad$^2$  \ \ \ \ \ 
Yunmo Chen$^2$\footnotemark[2]
} \\
$^1$University of California Los Angeles \quad \quad \quad \quad $^2$Bloomberg \\
{\small\texttt{tparekh@cs.ucla.edu, skothur@bloomberg.net}} \\
}

\begin{document}
\maketitle

\newcommand{\mypar}[1]{\vspace{0.35em}\noindent\textbf{#1}}
\newcommand{\SideNote}[2]{\todo[color=#1,size=\small]{#2}} 

\newcommand{\tanmay}[1]{\SideNote{orange!40}{#1 --tanmay}}
\newcommand{\ella}[1]{\SideNote{blue!40}{#1 --ella}}
\newcommand{\yunmo}[1]{\SideNote{brown!40}{#1 --yunmo}}

\newcommand{\modelName}{\textsc{PExA}}
\newcommand{\dataName}{Spider~2.0}

\newtcolorbox{promptbox}[1][]{
  colback=gray!5!white,
  colframe=gray!50!black,
  fonttitle=\bfseries,
  title=#1,
  coltitle=black,
  left=2mm, right=2mm, top=1mm, bottom=1mm,
  boxrule=0.5pt,
  sharp corners,
}

\newcommand\blfootnote[1]{%
  \begingroup
  \renewcommand\thefootnote{}\footnote{#1}%
  \addtocounter{footnote}{-1}%
  \endgroup
}

\newcommand{\red}[1]{\textcolor{red}{#1}}
\newcommand{\green}[1]{\textcolor{teal}{#1}}

\begin{abstract}
LLM-based agents for text-to-SQL often struggle with a latency-performance trade-off, where performance improvements come at the cost of latency or vice versa.
We cast text-to-SQL generation as a problem of passing software test coverage. Similar to its counterpart in software engineering, the original query is prepared with a suite of test cases with simpler, atomic  SQL queries that are executed in parallel and together ensure semantic coverage of the original query. After iterating on test case coverage, the final SQL is generated only when enough information is gathered, using the explored test case SQL queries to ground the final generation.
We validated our framework on a state-of-the-art benchmark for text-to-SQL, \dataName~\citep{spider2}, achieving a new state-of-the-art accuracy of  70.2\%.
\footnote{\label{note1}~At the time of submission on November 15, 2025.}
\end{abstract}

\section{Introduction}
\label{sec:intro}

Natural-language (NL) interfaces to databases unlock enterprise analytics for non-coding experts by translating NL questions into executable SQL queries \citep{text2sql_original, yu-etal-2018-spider}. While prior text-to-SQL work has shown strong performance on simpler academic benchmarks like SPIDER \cite{yu-etal-2018-spider} and BIRD \cite{bird}, these often do not capture the full complexity of real-world use. To evaluate text-to-SQL in more realistic settings, \citet{spider2} introduced \dataName, presenting significant challenges such as large cross-domain databases, nested data types, and long, complex SQL queries.

To address these complex challenges, researchers have increasingly turned to tool-augmented LLMs \cite{webgpt, toolformer}, deep planning \cite{cot, deepseek}, and agentic workflows \cite{react, reflexion}. Recent work in text-to-SQL has also explored such frameworks \cite{mag-sql, spider2}. However, these advanced methods face a well-known latency-performance trade-off \cite{scaling1, scaling2}. Performance improvements are often achieved through complex reasoning, multiple tool calls, and iterative self-correction, all of which substantially increase latency, making them less suitable for interactive analytics.

To achieve better Pareto optimality in this trade-off, we propose a novel framework that \emph{reframes the text-to-SQL task from a software testing perspective}. Rather than treating the NL query as a single complex problem to be solved sequentially, we treat it instead as a set of semantic requirements that must be ``covered.'' The agent's goal is to first generate a set of robust ``test cases'' -- simpler, self-contained SQL queries -- that collectively  cover the full semantics of the user's question and validate against the database. To boost coverage, test cases are intentionally \emph{over-generated} so that target and non-target database information can both be explored to guide the final SQL generation.

This testing paradigm is effective because it naturally enables parallelism.
Rather than a sequential chain of thought, we introduce diverse planning, concurrent execution, and single-step multi-path search, which enables parallel planning and execution of test case SQL.
This parallel exploration facilitates the gathering of all necessary information (e.g., table structures, value distributions, intermediate results) simultaneously and flattens the latency curve by bounding latency by the slowest operation rather than the sum of all operations.
Furthermore, it also improves performance by broader and more efficient semantic search.


To operationalize this paradigm, we develop \modelName, an agent comprising three specialized sub-agents: (1) \emph{Planner}: Decomposes the original problem into a set of self-contained, simpler questions for test case generation. (2) \emph{Test Case Generator}: Builds, revises, and executes the corresponding SQL test cases against the database in parallel to gather information. (3) \emph{Proposer}: Integrates the information gathered from the test cases to generate the final, long-form SQL for the original user query.

\begin{figure*}[t]
    \centering
    \includegraphics[width=0.95\linewidth]{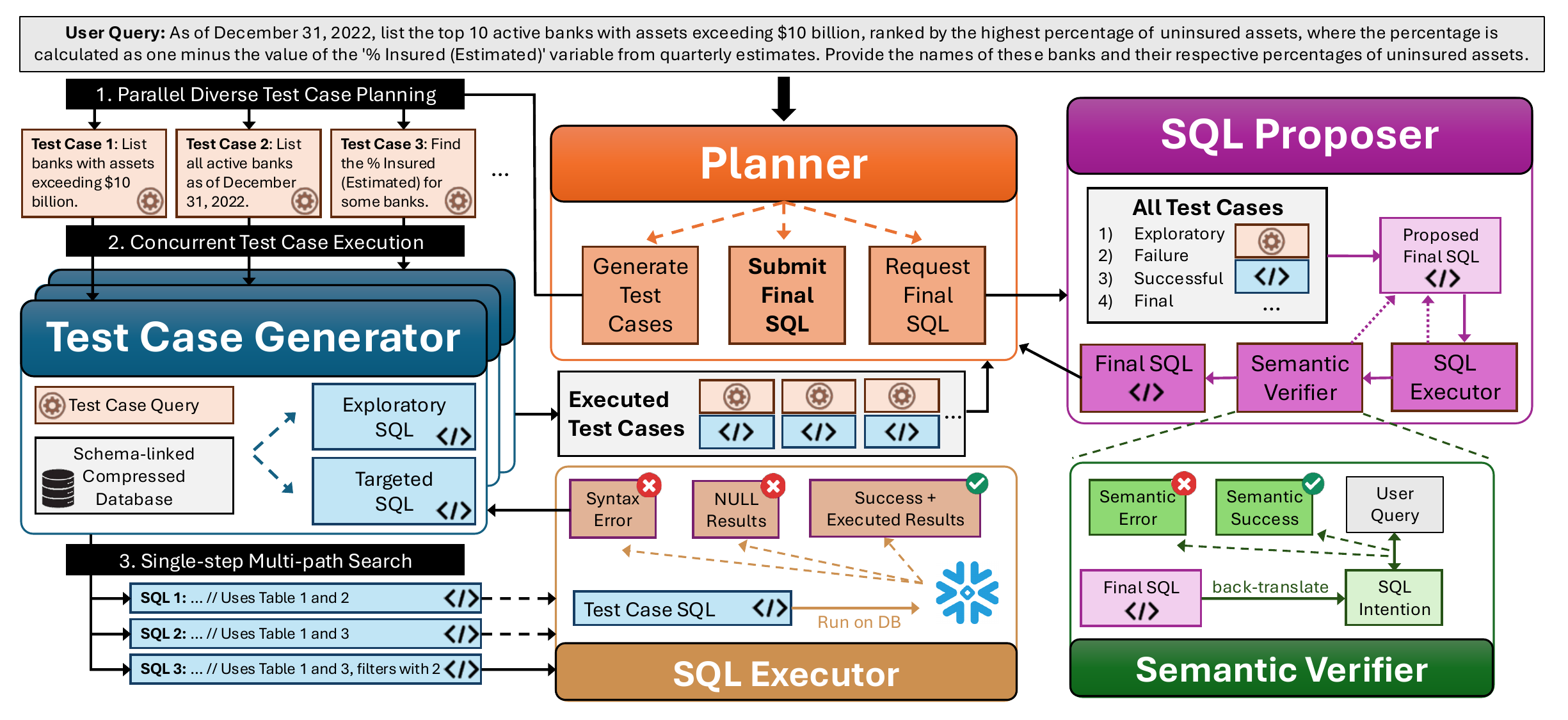}
    \caption{Illustration of our framework \modelName, comprising three sub-agents -- Planner, Test Case Generator, and SQL Proposer -- and two tools -- SQL Executor and Semantic Verifier. In the black boxes (left), we indicate how we induce parallelism in our framework. We indicate deterministic and agentic decision choices in terms of flow using solid and dotted arrows respectively.}
    \label{fig:workflow}
\end{figure*}

Our contributions include:
\begin{itemize}\setlength\itemsep{0em}\setlength\parskip{0pt}
    \item We reformulate the text-to-SQL task within the context of software testing and introduce parallelism to achieve better Pareto optimality between the latency and performance;
    \item We develop an agent, \modelName, to operationalize our proposed formulation and validate it by achieving the state-of-the-art\footref{note1} on the complex \dataName~benchmark.
\end{itemize}


\section{Related Work}
\label{sec:rel_works}

Here, we provide a brief overview of related work, with additional details in Appendix~\ref{sec:addn-rel-works}.

\paragraph{Text-to-SQL}
Spider 1.0 \cite{yu-etal-2018-spider} and BIRD \cite{bird} serve as foundational datasets for text-to-SQL, whereas \dataName{} \cite{spider2} is a recent, more challenging dataset to better evaluate methods on real-world cases.
While initial works focused on fine-tuning \cite{scholak-etal-2021-picard, codes}, advances in LLMs introduced better prompting-based methods \cite{din-sql}.
Recent frameworks also explore agents \cite{mag-sql, wang-etal-2025-mac} and inference-time scaling \cite{lee-etal-2025-mcs, reforce} to tackle complex text-to-SQL.
Similarly, we focus on building an agent but use test cases as a way to interact with databases.

\paragraph{Latency-Performance Trade-off}
Prior work has explored simple task decomposition \cite{least2most, decomposed} and sequential reasoning \cite{cot, tot, lats} to improve model performance. However, they require sequential execution and can increase latency. Task decomposition is also limited to explore only sub-tasks within  the original query, whereas our test cases explore a wider surface of the database robustly and provide the final SQL generation with extensive grounded context.
Inference-time scaling \cite{self-consistency, bon} and breadth-search \cite{saha-etal-2024-branch, dppm} enable multi-path search, but provide limited control over directions and thus are not efficient.
Our parallel exploration offers simple single-step parallel execution and a better guided multi-path search, providing an efficient, low-latency setting without degrading final SQL generation quality.

\paragraph{Software Testing}
Robust systems are built not by solving the full complexity of a program at once, but by ensuring that its behavior satisfies a broad suite of targeted tests \citep{software-testing, pressman2005software}.
The philosophy to ``cover'' specific functional requirements lies at the core of modern testing and test-driven development practices \citep{refactoring, beck2003test}.

\section{Approach}
\label{sec:methodology}


\subsection{A Software Testing Perspective}
\label{sec:components}

Inspired by software testing, we formulate text-to-SQL as a test coverage problem: \emph{creating a set of simpler, self-contained unit test cases that collectively ``cover'' the semantics of the user query and validate against the database}.
Different from query decomposition \citep[\emph{inter alia}]{reforce} where semantics present in sub-queries always reside in the original query, our testing paradigm covers a larger semantic surface via tests that extend beyond the original query to derive and explore additional information present in the database.
We encode this philosophy in \modelName\ as described in \autoref{fig:workflow}.


\subsubsection{Sub-Agents}

\paragraph{Planner}
The Planner implements the \textit{Test Planning and Modularization} principle in software testing \cite{software1}, determining the different semantic requirements of the user query and surrounding it with a suite of simpler verifiable (unit) test cases. 
In our framework, we define a test case as a self-contained SQL query that validates semantics present in or related to the user query -- such as a particular filter, join, aggregation, or existence check -- and provides independent and diverse evidence toward reconstructing the full SQL for the user query. 

In addition, the Planner serves as the central control module, orchestrating the interaction between the Test Case Generator and the SQL Proposer.
Such interactions are performed through test cases and their execution results.
The Planner determines when to continue exploratory test-case generation, and when sufficient evidence has been accumulated to invoke the Proposer for final SQL synthesis.
It also performs a final verification step, assessing the Proposer sub-agent's output before returning the final SQL and terminating.

\paragraph{Test Case Generator}
The Test Case Generator sub-agent corresponds to the \textit{Test Execution} phase in software testing \cite{test-execution}, where individual test cases are instantiated and run to gather empirical evidence.
The Test Case Generator translates each Planner-issued NL test case into an executable SQL query.
Given the potentially large and heterogeneous database, we follow ReFoRCE \cite{reforce} to perform lightweight LLM-based schema linking and database compression, identifying the relevant tables/columns required for the main user query with a focus on recall.
Conditioned on this pruned schema context, the sub-agent then synthesizes a precise SQL for each test case.
Importantly, we link the sub-agent with the \emph{SQL Executor} tool to build a feedback mechanism, verifying that the generated SQL executes successfully and entails the NL test case.

\paragraph{SQL Proposer}
The SQL Proposer sub-agent aligns with the \textit{Test Integration} stage in software testing \cite{test-execution}, where the results of multiple unit test cases are consolidated to derive the system’s overall intended behavior.
In our framework, given the executed test cases and the original user query, the SQL Proposer synthesizes the final long-form SQL by combining the partial semantic signals surfaced through the test-case results.
To maintain a focused and efficient context, the Proposer is intentionally not supplied with full database metadata and instead relies \emph{solely on the distilled evidence provided by the test cases}.
The module is further equipped with the same \emph{SQL Executor} and a \emph{Semantic Verifier} to ensure both valid execution and semantic fidelity before returning the final SQL.

\subsubsection{Tools}


\paragraph{SQL Executor}
The SQL Executor tool uses a SQL execution engine to execute the query on the database.
Based on the execution, it provides three kinds of feedback:
(1) Compilation Error, if there is a syntax issue,
(2) NULL Error, if the SQL executes but produces zero results, and
(3) Success, if the SQL executes with non-zero results, along with the truncated executed results.

\paragraph{Semantic Verifier}
This tool utilizes an LLM to ensure alignment between the NL instruction and generated SQL.
To do this, the proposed SQL is first back-translated into natural language, and then compared to the original user query to identify any semantic disparities.
If a huge disparity is found, the tool returns a semantic error.

\subsection{Parallel Exploration}
\label{sec:exploration}

Another major advantage of adopting a software testing perspective is that test cases are \emph{inherently parallelizable}. 
We exploit such parallel exploration in \modelName{} to boost performance and latency in three different ways, as described below.

\paragraph{Diverse Test Case Planning}
Our Planner prepares the user query with a suite of independent, self-contained test cases, running the entire planning phase in a single forward pass.
This parallel testing plan not only lowers planning overhead but also yields a more diverse set of semantic test cases, improving downstream coverage.

\paragraph{Parallel Test Case Execution}
Since the test cases are designed to not interact with each other, Test Case Generator can run all of them in parallel.
This parallel execution increases semantic breadth without scaling end-to-end latency, analogous to running a fully parallel test suite rather than a sequential battery of checks.

\paragraph{Single-step Multi-path Search}
For each test case being generated, we prompt \modelName{} to produce multiple candidate SQL queries with diverse reasoning strategies in one LLM generation call, enabling a structured multi-path search without sequential calls.
This single-step branching expands the search frontier early, allowing rapid pruning of weak hypotheses and more efficient exploration compared to sequential multi-solution exploration \cite{tot}.
This parallelization strategy bounds the time complexity to the slowest explored path, rather than the sum of all paths in an iterative framework.

\section{Experiments and Results}
\label{sec:expt}

\paragraph{Dataset}
We evaluate on the complex text-to-SQL \dataName{} benchmark, specifically on the Snow and Lite* versions.
\dataName-Snow comprises 547 examples that span 150+ databases, with an average of 800 columns per database in the Snowflake dialect.
\dataName-Lite* is a Lite version of the dataset, excluding the examples from BigQuery.\footnote{~Due to legal and cost constraints, we exclude BigQuery execution from our experiments.}

\paragraph{Evaluation Metrics}
We follow \citet{yu-etal-2018-spider} to consider Execution Accuracy (EX) and EX@4 (pass @ k) to measure model performance.\footnote{~To improve accuracy and stability in evaluations, the gold outputs for this dataset were updated on October 29, 2025. In this case, \emph{exact numbers of metrics might differ when compared to prior work}.
As most of our experiments were conducted prior to this update, we report our ablation and analytical results under the old evaluation setup. However, to be noted, our state-of-the-art performance at the time of submission was based on the updated evaluation data.}
For latency measurement, we report average wall time per user query.
GPT-o3 is used in all major experiments in this paper.

\paragraph{Baselines}
For baselines, we consider other open-source work on the \dataName{} leaderboard, specifically:
(1) Spider-Agent \citep{spider2}, a bash-based LLM agent,
(2) ReFoRCE \citep{reforce}, a database compression and inference-time scaling method,
(3) Chat2DB-Agent and AgenticData \cite{agenticdata}\footnote{~As the complete codebase is not available at the time of submission, we could not reproduce the results for comparison. Hence, we took the numbers directly from the leaderboard.}.

\begin{table}[t]
    \centering
    \small
    \renewcommand{\arraystretch}{1.25} 
    
    \begin{threeparttable}        
        \begin{tabular}{lccccc}
            \toprule
            \multirow{2}{*}{\textbf{Method}} & \multicolumn{2}{c}{\textbf{Snow}} & \multicolumn{2}{c}{\textbf{Lite}\tnote{*}} & \textbf{Wall} \\ 
            \cmidrule(lr){2-3} \cmidrule(lr){4-5}
             & \textbf{EX} & \textbf{EX@4} & \textbf{EX} & \textbf{EX@4} & \textbf{Time} \\ 
            \midrule
            Spider-Agent & 25.2 & 27.4 & 26.2 & 28.7 & 5.90 \\
            ReFoRCE      & 36.6 & 39.7 & 36.2 & 39.5 & \textbf{5.44} \\
            Chat2DB      & 44.1\tnote{1} & -- & -- & -- & -- \\
            AgenticData  & -- & -- & 44.5\tnote{1,2} & -- & -- \\ 
            \midrule
            \rowcolor{gray!10} 
            \textbf{\modelName{} (ours)} & \textbf{45.7} & \textbf{49.5} & \textbf{46.6} & \textbf{49.9} & \underline{5.55} \\ 
            \bottomrule
        \end{tabular}
        
        \begin{tablenotes}
            \footnotesize
            \item[1] As reported in the older leaderboard.
            \item[2] On complete Lite dataset.
        \end{tablenotes}
        \caption{Main results comparing the performance (EX, EX@4) and wall time (in mins) for \textbf{\modelName{}} against open-source baselines.}
        \label{tab:main-results}
    \end{threeparttable}
\end{table}

\subsection{Main Results}
\label{sec:results}

We present our main results in Table~\ref{tab:main-results}.
\modelName\ utilizes the inference-time scaling with majority consensus over four runs and outperforms all other baselines.
On the updated evaluation data, \textbf{\modelName{} achieves a new state-of-the-art \footref{note1} of 70.2\%} at the time of submission as shown in \S~\ref{sec:leaderboard}.
Meanwhile, parallel exploration ensures efficient and comparable wall time for \modelName.
We further provide a detailed theoretical and empirical study between sequential and parallel execution in Appendix~\ref{sec:latency-analysis}.

\begin{table}[t]
    \centering
    \small
    \renewcommand{\arraystretch}{1.2} 
    \begin{tabular}{lrr} 
        \toprule
        \textbf{Setting} & \textbf{EX} & \textbf{$\Delta$} \\
        \midrule
        \textbf{\modelName{} (Full)} & \textbf{42.9} & -- \\
        \midrule
        \hspace{2mm} w/o Plan-time parallelization & 40.0 & -2.9 \\
        \hspace{2mm} w/o Test-time parallelization & 39.9 & -3.0 \\
        \hspace{2mm} w/o Semantic Verifier         & 42.3 & -0.6 \\
        \hspace{2mm} w/o Proposer                  & 41.1 & -1.8 \\
        \bottomrule
    \end{tabular}
    \caption{Ablation study showing the performance impact when removing specific components from \modelName.}
    \label{tab:component-ablation}
\end{table}

\subsection{Analyses}
\label{sec:analysis}

For the various analyses, we consider a single run of \modelName{} on the Snow dataset, rather than the inference-time scaling performance (as reported in Table~\ref{tab:main-results}).
Additional analyses and results in \ref{sec:additional-results}.

\paragraph{Component Ablation}
We ablate various components used by our framework in \autoref{tab:component-ablation} (more details about model setup explained in Appendix~\ref{sec:appendix-component-ablation}).
Parallelization components are the most significant ones towards performance, while the semantic verifier and proposer also contribute to performance to some extent.
This analysis also demonstrates how the parallelization branching factor can be controlled for faster inference and lower model costs while incurring merely 2-3\% performance drop.

\begin{table}[t]
    \centering
    \small
    \renewcommand{\arraystretch}{1.25}
    \setlength{\tabcolsep}{8pt}
    \begin{tabular}{lllc}
        \toprule
        \textbf{Plan} & \textbf{Generate} & \textbf{Propose} & \textbf{EX} \\
        \midrule
        \multicolumn{4}{l}{\textit{Homogeneous models}} \\
        \multicolumn{3}{c}{GPT-o3}       & 42 \\
        \multicolumn{3}{c}{GPT-5}    & 45 \\
        \multicolumn{3}{c}{Sonnet-4} & 38 \\
        \multicolumn{3}{c}{Opus-4}   & 38 \\
        \midrule
        \multicolumn{4}{l}{\textit{Mixed models (Ablation)}} \\
        Sonnet-4     & GPT-o3       & GPT-o3       & 44 \\
        GPT-o3       & Sonnet-4     & GPT-o3       & 40 \\
        GPT-o3       & GPT-o3       & Sonnet-4     & 39 \\
        \bottomrule
    \end{tabular}
    \caption{Performance analysis comparing homogeneous against mix-and-match configurations on 100 samples from the \dataName{} dataset.}
    \label{tab:mix-and-match}
\end{table}

\paragraph{Studying LLM Synergies}
The components used by our agent can utilize different LLMs, enabling ablations of different LLMs and their synergies when used together.
In \autoref{tab:mix-and-match}, we study such effects with four LLMs.\footnote{Owing to higher costs of running frontier LLMs, we conduct this experiment on 100 samples from the dataset}
With our agent setup, we found that GPT-o3 and GPT-5 perform the best, while Claude models are slightly underperforming.
Component-wise ablations with Claude Sonnet-4 over GPT-o3 baseline revealed that Claude performs well on planning, while being poor at long-context/horizon reasoning for SQL generation.
The ability to combine different models allows a fine-grained control for our agent to achieve a better balance between performance and latency.

\begin{figure}[t]
    \centering
    \includegraphics[width=0.9\linewidth]{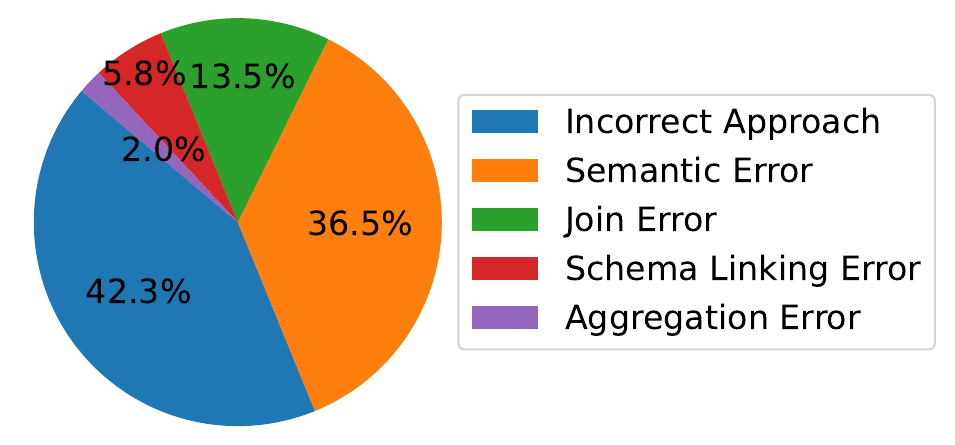}
    \caption{Error analysis and categorization on the major error categories from \modelName.}
    \label{fig:error-analysis}
\end{figure}

\begin{table}[t]
    \centering
    \small
    \renewcommand{\arraystretch}{1.25}
    \setlength{\tabcolsep}{10pt} 
    \begin{tabular}{c c | c c c} 
        \toprule
        \multicolumn{2}{c|}{\multirow{2}{*}{\textbf{Settings}}} & \multicolumn{3}{c}{\textbf{Execution Branches}} \\
        \cmidrule(lr){3-5}
        \multicolumn{2}{c|}{} & \textbf{1} & \textbf{2} & \boldmath$\infty$ \\
        \midrule
        \multirow{3}{*}{\shortstack{\textbf{Plan}\\\textbf{Branches}}} 
        & \textbf{1}        & 38.4 & 39.1 & 40.0 \\
        & \textbf{2}        & 38.9 & 39.5 & 41.1 \\
        & \boldmath$\infty$ & 39.9 & 41.6 & \textbf{42.9} \\
        \bottomrule
    \end{tabular}
    \caption{Performance analysis varying the parallelization degree. $\infty$ indicates no limit enforced.}
    \label{tab:branching-control}
\end{table}

\paragraph{Controlling the Efficiency-Performance Trade-off} \modelName{} offers granular control over the trade-off between computational cost and performance. While our default configuration leaves the search space unconstrained, we can restrict the branching factor during both the planning and execution phases to improve efficiency. \autoref{tab:branching-control} quantifies this relationship. We observe a 4.5\% drop in execution accuracy when limiting the system to a single branch compared to the unconstrained baseline. Consequently, increasing the branching factor consistently improves performance by enabling broader search coverage.

\paragraph{Error Analysis} We analyzed a sample of failure cases by comparing generated queries against gold reference SQL queries shown in \autoref{fig:error-analysis}. The breakdown reveals that semantic misinterpretations and flaws in initial planning/approaches are the dominant error sources, rather than simple syntactic mistakes. These high-level reasoning failures represent the primary bottleneck and suggest that future iterations should prioritize improved question comprehension and improve search in plans.

\section{Conclusion and Future Work}
\label{sec:conclusion}

In this work, we present \modelName, a text-to-SQL agent composed of a Planner, Test Case Generator, and SQL Proposer sub-agents. By utilizing parallelized exploration, \modelName{} efficiently traverses the search space even under strict latency constraints. Empirically, we demonstrate that our approach achieved state-of-the-art results on \dataName{} while maintaining wall-time comparable to existing baselines.
Future works can explore adapting our parallelized agentic architecture for wider benchmarking on other Text-to-SQL datasets, as well as broader software engineering and code generation tasks. 

\section*{Acknowledgements}

Tanmay Parekh would like to express his gratitude to Bloomberg for their support through the Bloomberg Data Science Ph.D. Fellowship.

\section*{Limitations}

Our baselines represent the state-of-the-art as of November 2025. While newer approaches may have emerged since, our baseline analysis is constrained as these methods are not open, preventing fair benchmarking.
Apart, we do not conduct any post-training while developing our framework.
While post-training can improve the agentic capabilities of the base LLM, we believe it can hamper the robustness.
Future works can explore better post-training with \modelName.

\section*{Ethical Considerations}

Our framework utilizes state-of-the-art LLMs like GPT-o3 off-the-shelf, and we do not filter/check for bias mitigation while using these LLMs. For practical utility, it is advised to consider separate bias mitigation strategies.
Also, since our framework is agentic in nature, it generates many LLM calls for the various agentic actions. Thus, running \modelName{} can be computationally expensive and should be considered accordingly.
We also acknowledge the utilization of LLMs/AI for paper writing and coding.

\bibliography{final_bib}

\clearpage

\appendix


\section{Additional Related Work}
\label{sec:addn-rel-works}

We provide a broader related work and the relative position of our work here.

\paragraph{Datasets for Text-to-SQL}
One of the prominent datasets utilized for Text-to-SQL is Spider 1.0 \cite{yu-etal-2018-spider}.
Subsequent works also created various variants like Spider-Realistic \cite{deng-etal-2021-structure}, Spider-DK \cite{gan-etal-2021-exploring}, and Spider-SS \cite{gan-etal-2022-measuring}.
To make the datasets more realistic and domain-specific, various new datasets like BIRD \cite{bird}, BookSQL \cite{kumar-etal-2024-booksql}, and MultiSQL \cite{li-etal-2024-multisql}.
Some works have also focused on specific domains with datasets like MIMICSQL \cite{mimicsql}, BioMedSQL \cite{biomedsql}, and LogicCat \cite{logiccat}.
To combine various challenges and create a complex industry-scale dataset, \dataName \cite{spider2} was proposed, and we evaluate our framework on this dataset in our work.

\paragraph{Automating Text-to-SQL}
Pre-LLM works \cite{wang-etal-2020-rat, lin-etal-2020-bridging, yin-etal-2020-tabert, arcadinho-etal-2022-t5ql} majorly relied on fine-tuning pre-trained language models on a set of training data.
The advent of LLMs introduced basic constraint-decoding and prompting methods \cite{scholak-etal-2021-picard, codexdb, chatgpt-eval}.
Some works focused on improving in-context examples \cite{din-sql, nan-etal-2023-enhancing, wang-etal-2024-improving-demonstration}, utilizing retrieval-augmented generation \cite{rag-gpt, chess, shen-etal-2024-improving}, and inference-time scaling \cite{lee-etal-2025-mcs, reforce, chasesql}.
With more compute, some works also show the effectiveness of fine-tuning small and large LLMs for text-to-SQL \cite{pourreza-rafiei-2024-dts, dubosql, he-etal-2025-star, arctic}.
Recent works have also proposed the use of multi-agent frameworks \cite{mag-sql, sqlfixagent, tool-assisted-agent, wang-etal-2025-mac} to tackle complex SQL queries.
Utilization of semantic verifiers \citet{sawhney-etal-2025-iterative} have also been explored lately.
In our work, we work on similar directions by proposing a simple multi-agent framework \modelName{} for the \dataName{} benchmark.

\paragraph{Analogy to Software Testing}
Software testing provides the conceptual backbone for the formulation of our agentic framework \modelName.
Classical works formalize software-testing as specification, coverage, and evidence aggregation rather than exhaustive verification, and emphasize designing suites of targeted tests to characterize system behavior \cite{software-testing, pressman2005software}.
Subsequent works have unified these ideas into rigorous coverage criteria, automated test execution, and oracle-based evaluation for determining correctness from multiple test outcomes \cite{test-execution}.
Our work also draws inspiration from the concept of executable specifications, central to test-driven development and agile engineering \cite{beck2003test}.
Structural and data-flow testing works also show how carefully chosen, small test units provide maximal information about program semantics \cite{selection-software}.
Overall, we root our conceptual design choices deeply in software testing.

\paragraph{Parallelization in LLMs}
Decomposition in planning \cite{least2most, decomposed, wang-etal-2023-plan} improves model performance; however, the co-dependent plans demand sequential execution and can increase the latency.
Newer approaches \cite{saha-etal-2024-branch, dppm} have explored parallel execution-based planning, however only in limited settings like evaluation and constrained generation where subtasks are more explicit.
Other works \cite{parekh-etal-2025-dynamic, DBLP:journals/corr/abs-2505-19435} explore dynamic planning of model reasoning methods to improve search.
Sequential reasoning \cite{cot} and various multi-path breadth search approaches \cite{tot, got, lats} iteratively explore each node, which increases latency.
On the other side, inference-time scaling approaches \cite{self-consistency, bon} that utilize temperature for multi-path search offer little control over the diversity in approaches, rendering them inefficient.
In our work, we propose parallel exploration, which offers the benefits of simple single-step parallel executable decomposition and a better guided multi-path search in an efficient, low-latency setting.

\section{Implementation Details}

Here, we describe the details of our implementation, including the prompts used for the different agents and the hyperparameters.

\subsection{\modelName{} Prompt Details}

Our \modelName{} pipeline comprises three major agents with distinctive roles. Similar to other unconstraining works \cite{parekh-etal-2025-dicore}, one of the advantages of our division is focused tasks for each agent and the reduction of constraints in the model prompts. We discuss some distinctions and complete prompts for each agent/component below.

\begin{figure*}[t]
\centering
\begin{tcolorbox}[fontupper=\small]
You are an excellent SQL agent whose task is to manage complex SQL query writing for a given natural language question. \\[0.5em]

\textbf{Tools Available to Agent:}

- \underline{Database Exploration}: Call the worker (generate\_testcase tool) to generate SQL queries for probes. Probes are simple natural language queries to provide basic information about the data to the proposer for writing the final SQL.

- \underline{Propose Final SQL}: Call the proposer (propose\_final\_sql\ tool) to integrate probes and synthesize the final SQL query. You should call the proposer only when you feel you have enough probes to generate the final SQL.

- \underline{Finish}: If you are confident that the final SQL query is ready and the executed results look good, you can submit it (finish tool). Else you can generate more probes to guide the final SQL generation. \\[0.5em]

\textbf{Guidelines:}

- Keep your probes simple and focused on exploring the data. They should be answerable by simple SQL queries.

- Make sure each probe is independent and self-contained and does not depend on previous probes. Do not include any probes for combining results or just for an extra
computation/filtering operation.

- Generate only as many probes as needed. Do not be redundant.

... \\[0.5em]

\textbf{Failure Examples:}

\underline{Question}: "Suppose it is 2020 currently. For each customer who purchased on the website in December last year, find the days between their first visit in December and their first purchase in December. Also find what type of device did they make that first purchase?"

\underline{Probes}: ["Retrieve a table of all customers who made purchases in December last year", "Retrieve a table of all customers who visited the website who also made purchases in December last year", "Retrieve a table of devices used by customers for their purchases. Assume there is a table which provides the customer id and device id and the purchase date", "Calculate the days between the first visit and the first purchase for customers in December last year"]

\underline{Explanation}:  The first probe does not specify the year. The second probe is not
simplest as it combines two different aspects (visits and purchases) into one unit.
The third probe assumes the existence of a table that may not be present in the
database. The fourth probe combines probes 1 and 2 to calculate the days - thus, not
self-contained. You should not write probes like the fourth.

... \\[0.5em]

\textbf{Correct Examples:}

\underline{Question}: "Suppose it is 2020 currently. For each customer who purchased on the website in December last year, find the days between their first visit in December and their first purchase in December. Also find what type of device did they make that first purchase?"

\underline{Probes}: ["Retrieve a table of all customers who made purchases in December last year (the current year is 2020)", "Retrieve a table of all customers who visited the website in December last year (the current year is 2020)", "Retrieve a table of devices used by customers for their purchases"]

... \\[0.5em]

\textbf{Test Data}

\underline{Question}: As of December 31, 2022, list the top 10 active banks with assets exceeding \$10 billion, ranked by the highest percentage of uninsured assets, where the percentage is calculated as one minus the value of the `\% Insured (Estimated)' variable from quarterly estimates. Provide the names of these banks and their respective percentages of uninsured assets.

\underline{Another way to say it}: As of December 31, 2022, list the top 10 active banks with assets exceeding \$10 billion, ranked by the highest percentage of uninsured assets, where the percentage is calculated as one minus the value of the `\% Insured (Estimated)' variable from quarterly estimates. Provide the names of these banks and their respective percentages of uninsured assets.

Now based on the instructions and examples provided, generate the probes (using the
generate\_testcase tool) for the above question.

\end{tcolorbox}
\caption{Prompt utilized for \modelName's Planner}
\label{fig:planner-prompt}
\end{figure*}

\paragraph{Planner}
We provide the prompt utilized for the Planner in Figure~\ref{fig:planner-prompt}.
Specifically, we describe the tools available, provide extensive guidelines as well as both positive and negative in-context examples, and final test question information.
Note that the prompt is short and the LLM is encouraged to think and plan for the main question.

\begin{figure*}[t]
\centering
\begin{tcolorbox}[fontupper=\small]
You are an SQL expert. Your task is to generate as many diverse SQL queries as possible
for the question below. Each SQL should use a different combination of tables, so that at
least one will execute correctly. \\[0.5em]

\textbf{Instructions:}

1. \underline{Check Question Completeness:} If the question is not self-contained or unanswerable, use
the "not enough information" tool with an explanation.

2. \underline{Generate many Diverse SQL queries:} If answerable, generate as many diverse SQL queries as possible (even if you are confident about your first choice). Each SQL should use a different set of tables. Do NOT simply use the same table with different joins, filters, or aliases and call them diverse. True diversity means using fundamentally different tables or table combinations, not just different ways of querying the same table. For each SQL, provide a short description of what it does and how it is different from the others. Submit this list of SQL queries to the "execute\_sql" tool for execution validation.

3. \underline{Explore Table Schema:} Make sure that you are entirely sure about the attribute values before writing the SQL. If you are unsure, you can explore table columns or values using the "execute\_sql" tool (using regex or ILIKE functions). Set the exploration attribute to True for such queries. Batch and execute all such exploration queries in a single tool call. Do not explore tables one by one in separate tool calls.

4. \underline{Tool Usage:} In every response, you must use one of the tools. \\[0.5em]

\textbf{SQL Writing Guidelines:}

- MUST stay faithful to the exact wording of the question. Prefer exact table/column matches over partial matches. If an exact-match table exists, do not union with partial matches; only explore alternates if no exact match is found. Examples: If the question mentions `total functional expenses', only use tables which mention them exactly.

- MUST keep queries simple (Occam's razor). Avoid extra joins/filters/rankings unless explicitly requested. Examples: If the question asks for “average measurement in a month”, do not average across hours — average across days.

- MUST pay careful attention to any External Knowledge hints provided below; treat them as guidance for table/column.

- SHOULD prefer inner joins (which naturally drop NULLs) unless the question requires outer joins. In general, remove rows with NULLs unless told otherwise. ... \\[0.5em]

\textbf{Dialect, Naming, and Qualification}

- MUST use Snowflake SQL. Enclose all identifiers (databases, schemas, tables, columns, aliases) in double quotes.

- MUST fully qualify table names using the format: "<DATABASE>"."<SCHEMA>"."<TABLE>". To reference INFORMATION\_SCHEMA, use "<DATABASE>"."INFORMATION\_SCHEMA"."TABLES" instead of INFORMATION\_SCHEMA.TABLES or anything else.

 ... \\[0.5em]

\textbf{Examples}

\underline{Example 1: (Diverse SQL queries)}

Question: Retrieve the weather for all active days at an XYZ school for 2022.

Tool Call:
execute\_sql([
(`description': "Uses Table 1 and Table 2: Joins school active days with weather data.",
`sql\_query': "SELECT ...", `exploration': False),
(`description': "Uses Table 4 and Table 3: Alternative join for weather info.", `sql\_query':
"SELECT ...", `exploration': False)
])

... \\[0.5em]

\textbf{Table Info:}

\underline{Table full name:}

FINANCE\_ECONOMICS.CYBERSYN.BANK\_FOR\_INTERNATIONAL\_SETTLEMENTS\_ATTRIBUTES
Column name: COUNTERPARTY\_GEO\_NAME Type: TEXT Description: Country or country group, representing the counterparty geography for the variable

... \\[0.5em]

\textbf{Table Names:}

The table structure information is (\{database name: \{schema name: [table name]\}\}):

\{`FINANCE\_ECONOMICS': \{`CYBERSYN': ... \\[0.5em]

\textbf{External Knowledge:}
... \\[0.5em]

\textbf{Question:}
Retrieve a table of all banks with their identifiers, names, total assets, the `\% Insured (Estimated)' value, and the reporting date for the quarter ending `2022-12-31'.

\end{tcolorbox}
\caption{Prompt utilized for \modelName's Test Case Generator}
\label{fig:generator-prompt}
\end{figure*}

\paragraph{Test Case Generator}
We provide the prompt utilized for the Test Case Generator in Figure~\ref{fig:generator-prompt}.
We provide the main instructions first with detailed steps.
Then, we provide an extensive guidelines for writing SQL queries in the dialect, and domain-specific instructions.
We provide some examples for diverse SQL queries and batched exploration queries.
We then provide the table information with table names.
Finally, we provide any external knowledge if present and the main question/probe from the planner.
We ask the LLM to provide two kinds of SQL queries - one which answers the question, and the other which is used to explore the database for more information.
This is the only place where we provide the long table information in entirety and encourage the Generator LLM to utilize its long context reasoning.

\begin{figure*}[t]
\centering
\begin{tcolorbox}[fontupper=\small]
You are an SQL expert. Your task is to generate a final SQL query for the given question. You are provided with a set of probes and corresponding SQL queries to provide you information about the database structure. \\[0.5em]

\textbf{SQL Writing Guidelines:}

- MUST stay faithful to the exact wording of the question. Do not add extra tables/filters/grouping beyond what is asked. Examples: If the question asks about top rising terms, do not add top rising
international terms ... 

- MUST keep queries simple (Occam's razor). Avoid extra ranking (RANK = 1), FLOOR() on dates, redundant GROUP BY, or unnecessary unions unless explicitly requested. Examples: If the question asks for “overall score”, do not average “overall direct”
and “overall indirect”; use overall direct as overall score ...

- MUST try to be faithful to the Final Probes. Example: If the probe uses "DICOM\_ALL" and results are non-null, prefer "DICOM\_ALL" in the final SQL (not "DICOM\_PIVOT" or others) 

... \\[0.5em]

\textbf{Dialect, Naming, and Qualification}

- MUST use Snowflake SQL and enclose all identifiers (DBs, schemas, tables, columns, aliases) in double quotes.

- MUST fully qualify tables: "<DATABASE>"."<SCHEMA>"."<TABLE>". To reference
INFORMATION\_SCHEMA, use "<DATABASE>"."INFORMATION\_SCHEMA"."TABLES" instead of
INFORMATION\_SCHEMA.TABLES or anything else.

- Preserve output case (uppercase/lowercase/mixed) as stored; do not LOWER()/UPPER()
results. 

...  \\[0.5em]

\textbf{Here are the details:}

\underline{Main Question:} As of December 31, 2022, list the top 10 active banks with assets exceeding \$10 billion, ranked by the highest percentage of uninsured assets, where the percentage is calculated as one minus the value of the `\% Insured (Estimated)' variable from quarterly estimates. Provide the names of these banks and their respective percentages of uninsured assets.

\underline{Another way to say it:} As of December 31, 2022, list the top 10 active banks with assets exceeding \$10 billion, ranked by the highest percentage of uninsured assets, where the percentage is calculated as one minus the value of the `\% Insured (Estimated)' variable from quarterly estimates. Provide the names of these banks and their respective percentages of uninsured assets. \\[0.5em]

\textbf{Probes and SQL Queries:} \\[0.5em]

\underline{Exploratory Probes:}

1. Probe: Retrieve a count of banks that reported total assets exceeding 10,000,000,000 (ten billion) dollars for the quarter ending `2022-12-31'.

SQL: SELECT "VARIABLE", "VARIABLE\_NAME" FROM "FINANCE\_ECONOMICS"."CYBERSYN". "FINANCIAL\_INSTITUTION\_ATTRIBUTES" WHERE LOWER("VARIABLE\_NAME") LIKE `\%total asset\%';

Executed Result (truncated):

VARIABLE,VARIABLE\_NAME

ASSET5,Average Total Assets

ASSET,Total Assets

...\\[0.5em]

\underline{Failure Probes:}

...\\[0.5em]

\underline{Successful Probes:}

...\\[0.5em]

\underline{Final SQL queries (pay more attention here):}

...\\[0.5em]

Now use these information to generate a final SQL query that answers the question. The final SQL query should be a complete and valid SQL query that can be executed on the database.

\end{tcolorbox}
\caption{Prompt utilized for \modelName's Final SQL Proposer}
\label{fig:proposer-prompt}
\end{figure*}

\paragraph{SQL Proposer}
We provide the prompt utilized for the Test Case Generator in Figure~\ref{fig:proposer-prompt}.
For the SQL proposer, we start with similar SQL writing and domain-specific guidelines as for the Test Case Generator.
Next, we provide the main question.
After this, we provide all the probes from the test case generator, categorized into four clusters:
(1) Exploratory - These are used to explore the database and provide additional information.
(2) Failure - These SQL queries failed and provide useful information for how not to write the final SQL.
(3) Successful - These SQL queries succeeded and provide information for what are other ways to answer the question.
(4) Final SQL queries - These are the most important and chosen by the test case generator as the most promising SQL queries for answering the question.
Based on all this information, we ask the Proposer LLM to write the final SQL based on its long-horizon reasoning.

\begin{figure*}[t]
\centering
\begin{tcolorbox}[fontupper=\small]
You are a semantic verifier. Your task is to verify the semantic correctness of a SQL query against a natural language question. \\[0.5em]

\textbf{Here are the steps you should follow:}

1. \underline{Back-translate the SQL query:} Convert the SQL query (conditioned on the executed result) into natural language query that captures what the SQL query is trying to achieve. This should be a faithful representation of the SQL query and the executed result in natural language.

2. \underline{Comparison for Semantic Verification:} Compare the back-translated natural language query (from part 1) with the original question. The back-translated query should capture the same intent and meaning as the original question.

\quad - If the back-translated query matches the original question, it indicates that the SQL query is semantically correct.

\quad - If the back-translated query does not match the original question, it indicates that the SQL query is semantically incorrect.

3. \underline{Return a response:} Your final response should be a structured JSON of three fields - "correct", "explanation", and "back\_translated\_query". The "back\_translated\_query" should always be the query back-translated from step 1. Based on the comparison in
step 2:

\quad - If the SQL query is semantically correct: "correct" = true and "explanation" = <Provide a one line explanation of why the back-translated query matches the original question.>

\quad - If the SQL query is semantically incorrect: "correct" = false and "explanation" = "<Give one line explanation of what is the difference between the back-translated query. Add one line for what can be potential improvement to fix this issue.>" \\[0.5em]

\textbf{Original Question:} As of December 31, 2022, list the top 10 active banks with assets exceeding \$10 billion, ranked by the highest percentage of uninsured assets, where the percentage is calculated as one minus the value of the `\% Insured (Estimated)' variable from quarterly estimates. Provide the names of these banks and their respective percentages of uninsured assets. \\[0.5em]

\textbf{SQL Query:} WITH combined as SELECT ... \\[0.5em]

\textbf{Executed Result:}

Bank,Asset,\% Insured (Estimated), ... \\[0.5em]

\textbf{Some final guidelines for your semantic matching:}

- Be extra careful with complex regex and try to ensure they are written correctly. If not provide an explanation to improve the regex.

- Check the COALESCE statements carefully. Verify that they are not redundant the default values are not incorrect. We want to keep the queries simple and avoid any redundant/overthought statements.

- Check the filters carefully. Ensure that there are no redundant filters which are unrelated to the original question. For example, if the question asks for trees not marked as dead, the SQL should only have filters for status not `dead'. We want to avoid additional status checks like `stump', `cut down', `removed' etc.

... \\[0.5em]

Now please follow the steps above and return your response in the specified JSON
format.

\end{tcolorbox}
\caption{Prompt utilized for \modelName's Semantic Verifier}
\label{fig:semantic-verifier-prompt}
\end{figure*}

\paragraph{Semantic Verifier}
Finally, we provide the prompt utilized for the Semantic Verifier in Figure~\ref{fig:semantic-verifier-prompt}.
We provide the three major steps involved in semantic verification:
(1) Back-translation of the SQL query into natural language,
(2) Comparison of translated query and original question to find semantic intent, and
(3) Response generation with explanation.
We then provide the original question, the LLM-written SQL query, and the execution results for its reference.
Finally, we also provide some specific guidelines to steer its behavior.

\begin{table}[t]
    \centering
    \small
    \begin{tabular}{lc}
        \toprule
        \textbf{Hyperparameter} & \textbf{Value} \\
        \midrule
        Number of workers & 6 \\
        Max Planner Iterations & 20 \\
        Max Testcase Iterations & 15 \\
        Max Proposer Iterations & 20 \\
        \midrule
        LLM temperature & 0.3 \\
        Verifier temperature & 1.0 \\
        \midrule
        Requests per second & 0.5 \\
        Max retries & 15 \\
        Timeout & 120s \\
        \midrule
        Execution Format & csv \\
        Execution Max Length & 500 \\
        Execution Max Rows & 3 \\
        \bottomrule
    \end{tabular}
    \caption{Hyperparameter values for \modelName.}
    \label{tab:hyperparameters}
\end{table}

\subsection{Code Implementation}

We provide the hyperparameters set for the \modelName{} runs in Table~\ref{tab:hyperparameters}.
We use LangGraph\footnote{\url{https://www.langchain.com/langgraph}} for implementing our LLM agent.
To avoid endless LLM loops, we limit the max number of iterations per LLM agent (as highlighted in the table).
We utilize API from OpenAI and Anthropic for calling respective LLMs.

\section{Additional Results and Analyses}
\label{sec:additional-results}

Here we provide additional results and analyses to support our main results.

\subsection{Latency analysis for sequential vs. parallelization}
\label{sec:latency-analysis}

To objectively evaluate the speed gains achieved by our parallelization, we conduct a theoretical latency analysis for sequential vs. parallelization with \modelName.
First, we build a theoretical model for studying the latency of \modelName{} architecture. Specifically, we split the total latency into the latency of each component as
$$
    t_{total} = t_{plan} + t_{generate} + t_{propose}
$$
SQL Proposer operates the same way for both sequential and parallel execution and we will focus only on $t_{plan}$ and $t_{generate}$ for our analysis.

In the sequential case, the latency term of $t^s_{plan} + t^s_{generate}$ can be written as
$$
    t^s_{plan} + t^s_{generate} = \sum_{i=1}^{k_1} \left(t_{plan}^i + \sum_{j=1}^{k_{i2}} t_{generate}^{ij} \right)
$$
where $t_{plan}^i$ and $t_{generate}^{ij}$ indicates the time for making the $i$-th plan and executing it using the $j$-th strategy. Here $k_1$ is the total number of plans and $k_{i2}$ indicates the total execution depth tried for this $i$-th plan.

In a similar way, when run in parallel as in \modelName, in sequential this term can be written as
\begin{align*} 
    t^p_{plan} + t^p_{generate} &= max_{i=1}^{k_1} \quad t_{plan}^i \quad + \\
    & \quad \quad max_{i=1, j=1}^{i=k_1, j=k_{i2}} \quad t_{generate}^{ij}
\end{align*}

\begin{table}[t]
    \centering
    \small
    \begin{tabular}{lc}
        \toprule
        \textbf{Component} & \textbf{Average Latency (s)} \\
        \midrule
        Planner & 18.9 \\
        Test Case Generator & 20.4 \\
        SQL Executor & 2.4 \\
        SQL Proposer & 51.8 \\
        Semantic Verifier & 10.6 \\
        \bottomrule
    \end{tabular}
    \caption{Component-wise latency analysis using o3 as the LLM in \modelName.}
    \label{tab:component-latency}
\end{table}

\begin{table}[t]
    \centering
    \small
    \begin{tabular}{lc}
        \toprule
        \textbf{Mode} & \textbf{Total Latency (s)} \\
        \midrule
        Sequential & 680 \\
        Parallel & 351 \\
        \bottomrule
    \end{tabular}
    \caption{Theoretical model latencies for \modelName in sequential vs. parallel modes.}
    \label{tab:pexa-seq-vs-parallel}
\end{table}

Now, to objectify empirically, we measure the average latency of each component, as shown in Table~\ref{tab:component-latency}.
Using average conservative values for $k_1=3$ and $k_2$ sampled from normal distribution with $\mu=8$ and $\sigma=2$, we can estimate the time for the sequential and parallel cases of \modelName{} as shown in Table~\ref{tab:pexa-seq-vs-parallel}.
The above values are chosen based on empirical estimation of the actually observed ones.
For parallel case, we do account for communication delay with a liberal 5-20\% per max operation.

Overall, we note how parallel exploration, theoretically, brings about a latency improvement of nearly 2x relative to sequential exploration of the same search space.
In practical scenarios, this gain is slightly higher owing to lower communication delays and early stopping owing to the parallel search.

\begin{table}[t]
    \centering
    \small
    \begin{tabular}{l|ccc}
        \toprule
        \textbf{Iteration Number} & \textbf{SCR $\uparrow$} & \textbf{OR $\downarrow$} & \textbf{\# TC $\downarrow$} \\
        \midrule
        1.0 & 78-86\% & 25\% & 5.0 \\
        2.0 & 86-91\% & 24\% & 4.2 \\
        2.1 & 85-87\% & 23\% & 4.0 \\
        2.2 & 84-96\% & 20-21\% & 3.9 \\
        2.3 & 90-91\% & 15-17\% & 3.6 \\
        3.0 & 90-94\% & 13-16\% & 3.7 \\
        3.1 & 88-92\% & 13-22\% & 3.7 \\
        3.2 & 87-95\% & 13-21\% & 3.3 \\
        4.0 & 91-97\% & 15-21\% & 3.5 \\
        4.1 & 92-95\% & 13-14\% & 3.1 \\
        \bottomrule
    \end{tabular}
    \caption{Tracking self-containment rate (SCR), overlap rate (OR), and number of proposed test cases (\# TC) by the planner across iterations of prompt-based steering. The LLM used for Planner for these iterations was o3-mini and number of runs was 3 per iteration.}
    \label{tab:planner-iterative-improvement}
\end{table}

\subsection{Deep-dive into Planner}

In this analysis, we provide additional studies to highlight how our Planner agent operates.
While the larger goal of Planner is to reduce the user query into smaller test cases, it is more than a simple decomposition.
The generated test cases are aimed to "over-generate" and can have overlap across each other - but the main goal remains to semantically holistically cover the main user query.
To steer the model towards this desired behavior, we rely on two in-context examples and guidance-based prompting.
To verify the quality of the Planner across iterations, we track metrics like self-containment ratio, overlap ratio, and the number of proposed test cases.
Ideally, we want a high self-containment rate with the lowest overlap ratio and number of proposed test cases.
We measured the performance on these metrics utilizing Claude Sonnet-4.
We provide the iterative improvement of the Planner (each iteration is an improved prompt setup) in Table~\ref{tab:planner-iterative-improvement}.
Clearly, our guidance-based steering improved the quality of the generated plans by the Planner.
Some of the updated definitions of our test cases / probes are provided below for reference:
\begin{itemize}
    \item Probes are simple natural language queries to provide basic information about the data to the proposer for writing the final SQL
    \item Keep your probes simple and focused on exploring the data. They should be answerable by simple SQL queries
    \item Make sure each probe is independent and self-contained and does not depend on previous probes. Do not include any probes for combining results or just for an extra computation/filtering operation.
\end{itemize}

\subsection{Confidence Error Bounds}

Owing to the high cost of executing multiple runs end-to-end, we could not provide strong confidence bounds for all models and runs.
However, across four runs for \modelName{} and ReFoRCE, the confidence error bounds were 0.9\% and 1.1\% EX, respectively.
This indicates how the variation across runs for PExA is comparable to ReFoRCE and other previous works.

\subsection{Component-wise Ablations - Implementation Details}
\label{sec:appendix-component-ablation}

We discussed and provided component-wise ablations in \S~\ref{sec:analysis}.
Here, we provide additional implementation details for each ablation for additional information:
\begin{itemize}
    \item w/o Plan-time parallelization: removes the parallelization of generating self-sufficient independent plans from the Planner, which leads to sequential planning. The test case generator is also run one at a time for this version. The other components remain the same.
    \item w/o Test-time parallelization, removes the single-step multi-path search in the Test Case Generator. This leads to the exploration of one SQL at a time. The other components remain the same.
    \item w/o the Semantic Verifier, removes this tool from PExA. Thus, the agent only relies on self knowledge and syntactic verification for feedback on its SQL. The other components remain the same.
    \item w/o Proposer, removes the SQL Proposer and asks the Planner to directly propose the final SQL. The Planner still has access to the other tools for verification and feedback. The other components remain the same.
\end{itemize}

\subsection{Controlling efficiency with smaller LLMs}

\begin{table}[t]
    \centering
    \small
    \begin{tabular}{lll|cc}
        \toprule
        \textbf{Plan} & \textbf{Generate} & \textbf{Propose} & \textbf{EX} & \textbf{Wall Time}\\
        \midrule
        o3 & o3 & o3 & 42 & 5.55 \\
        o3-mini & o3 & o3 & 42 & 5.01 \\
        o3 & o3-mini & o3 & 41 & 3.95 \\
        o3 & o3 & o3-mini & 39 & 4.77 \\
        o3-mini & o3-mini & o3 & 32 & 3.47 \\
        o3-mini & o3 & o3-mini & 35 & 4.36 \\
        o3 & o3-mini & o3-mini & 31 & 3.50 \\
        o3-mini & o3-mini & o3-mini & 28 & 2.99 \\
        \bottomrule
    \end{tabular}
    \caption{Performance and latency (in mins) analysis for mix-and-match in \modelName{} with four different LLMs.}
    \label{tab:mix-and-match-efficiency}
\end{table}

Apart from controlling the branching factor, we provide another way to improve model efficiency - by replacing the LLMs for different sub-components with smaller ones.
To this end, we provide a comprehensive table of using a combination of o3-mini and o3 as the LLMs for the three different components of our pipeline on 100 datapoints from \dataName-Snow in Table~\ref{tab:mix-and-match-efficiency}.
As noted, replacing a single component in a pure o3 pipeline with o3-mini can improve the efficiency by 20\% while suffering a slight drop relatively.
On the other hand, if 2-3 components used are with smaller LLMs, then a more drastic performance drop can be noticed.
Overall, we believe, the Proposer and Test Case Generator are the major components that need a stronger LLM to do long-context and long-horizon reasoning.

\subsection{Characterization of \modelName{} runs}

\begin{table}[t]
    \centering
    \small
    \begin{tabular}{l|cc}
        \toprule
        \textbf{Statistic} & \textbf{Correct} & \textbf{Incorrect} \\
        \midrule
        \# Output Tokens & 19,631 & 29,312 \\
        \# Generation Tokens & 6,653 & 9,823 \\
        \# Reasoning Tokens & 12,977 & 19,488 \\
        \# LLM Calls & 16.74 & 22.63 \\
        \midrule
        \# Planner Generates Plans & 1.26 & 1.41 \\
        \# Parallel Plans & 7.06 & 7.2 \\
        \# Probes Generated & 29.2 & 41.9 \\
        \# SQL Executions & 8.99 & 13.48 \\
        \# Semantic Verifications & 1.09 & 1.32 \\
        \midrule
        Planner Recursion Reached & 0\% & 5\% \\
        Generator Recursion Reached & 2.8\% & 5.4\% \\
        Proposer Recursion Reached & 0.8\% & 6\% \\
        \bottomrule
    \end{tabular}
    \caption{Averaged statistics per datapoint for \modelName{} run on \dataName-Snow dataset using the o3 model as the LLM.}
    \label{tab:run-stats}
\end{table}

Here, we provide some basic statistics of \modelName{} using o3 on the \dataName-Snow dataset to characterize some insights about the model runs.
We split these statistics on successful/correct runs vs. incorrect ones and detail them in Table~\ref{tab:run-stats}.
First, we note how incorrect runs need 50\% more tokens and cost than correct ones. They also need 35\% more LLM calls.
This is mainly owing to the LLM Agent trying to find a solution or needing more reasoning steps to solve the problem at hand.
To improve efficiency for such runs, one can reduce the recursion limits for the different components.

At the same time, we note that the number of LLM calls by \modelName{} is on the higher end, indicating higher model costs.
In our best opinion, it is really difficult to balance and improve the Pareto optimality across all three of cost, latency, and performance.
In our work, we focus on improving the Pareto optimality for latency and performance, while our model costs do increase as an after-effect.

Next, we note that the number of planner calls and plans generated is in the range of 1.25-1.4 and 7-7.2, respectively, providing insights for the level of decomposition done for problem solving.
While planner statistics are similar for correct/incorrect runs, in terms of execution, incorrect runs are 45-50\% more expensive as measured by the number of probes and SQL executions.

Finally, we note the rate of recursion limit hits. For correct runs, it's low, touching about 3\%.
However, for incorrect runs, this number touches 5-6\% consistently for all agents in \modelName.
This indicates that a part of the failure of the incorrect runs lies in the limited reasoning steps and possibly can be improved by expending more resources.

\subsection{Experiments with Gold Schema}

\begin{table}[t]
    \centering
    \small
    \begin{tabular}{l|cc}
        \toprule
        \multirow{2}{*}{\textbf{Method}} & \multicolumn{2}{c}{\textbf{Snow}} \\
        & \textbf{EX} & \textbf{EX@4} \\
        \midrule
        \modelName & 45.7 & 49.5 \\
        \modelName{} w/ Gold Schema & 47.2 & 50.8 \\
        \bottomrule
    \end{tabular}
    \caption{Performance comparison for \modelName{} with and w/o Gold Schema.}
    \label{tab:gold-schema}
\end{table}

To study the impact of gold schema, and schema linking in general, we run \modelName{} with gold schema and compare it without the gold schema in Table~\ref{tab:gold-schema}.
We note how the improvement by using the gold schema for both metrics is minimal, capped at about 1.5\%.
This aligns with our error analysis as well in \S~\ref{sec:analysis}, and we conclude that schema linking by itself is not a major issue to improve performance.
Instead, we posit that logical reasoning and long-context reasoning are better improvement areas.

\begin{table*}[t]
    \centering
    \small
    \begin{tabular}{l|p{13.5cm}}
        \toprule
        \multicolumn{2}{c}{\textbf{Example 1}} \\
        \midrule
        \textbf{Question} & Which weather stations in Washington State recorded more than 150 rainy days in 2023 but fewer rainy days compared to 2022? Defining a “rainy day” as one having precipitation greater than zero millimeters and not equal to 99.99. Only include stations with valid precipitation data. \\

        \midrule
        
        \multirow{12}{*}{\textbf{Spider-Agent}} & \begin{lstlisting}[language=SQL]
WITH rainy_counts AS (  SELECT  "station_id",  DATE_PART('year', "date") AS "year",  COUNT_IF("precipitation" > 0) AS "rainy_days" FROM NOAA_DATA.NOAA_GSOD.GSOD_DAILY WHERE  "state" = 'WA'  AND DATE_PART('year', "date") IN (2022, 2023)  GROUP BY  "station_id", DATE_PART('year', "date") ), counts_2022 AS (  SELECT "station_id", "rainy_days"  FROM rainy_counts  WHERE "year" = 2022 ), counts_2023 AS (  SELECT "station_id", "rainy_days"  FROM rainy_counts  WHERE "year" = 2023 ) SELECT  s."station_id",  s."name" FROM counts_2023 c23 JOIN counts_2022 c22  ON c23."station_id" = c22."station_id" JOIN NOAA_DATA.NOAA_GSOD.STATIONS s  ON s."station_id" = c23."station_id" WHERE c23."rainy_days" > 150  AND c23."rainy_days" < c22."rainy_days";
        \end{lstlisting} \\

        \midrule

        \multirow{16}{*}{\textbf{ReFoRCE}} & \begin{lstlisting}[language=SQL]
WITH rainy_2023 AS ( SELECT  "stn",  "wban",  COUNT(*) AS "rainy_days_2023"  FROM NOAA_DATA.NOAA_GSOD."GSOD2023"  WHERE "prcp" IS NOT NULL AND "prcp" <> 99.99 AND "prcp" > 0 GROUP BY "stn", "wban" ), rainy_2022 AS ( SELECT  "stn",  "wban",  COUNT(*) AS "rainy_days_2022"  FROM NOAA_DATA.NOAA_GSOD."GSOD2022"  WHERE "prcp" IS NOT NULL  AND "prcp" <> 99.99  AND "prcp" > 0  GROUP BY "stn", "wban" ), counts AS ( SELECT  r23."stn",  r23."wban",  r23."rainy_days_2023",  r22."rainy_days_2022"  FROM rainy_2023 r23  INNER JOIN rainy_2022 r22  ON r23."stn" = r22."stn" AND r23."wban" = r22."wban" ) SELECT  s."usaf" AS "station_usaf",  s."wban" AS "station_wban",  TRIM(s."name") AS "station_name",  counts."rainy_days_2023",  counts."rainy_days_2022" FROM counts JOIN NOAA_DATA.NOAA_GSOD."STATIONS" s ON counts."stn" = s."usaf" AND counts."wban" = s."wban" WHERE s."state" = 'WA' AND counts."rainy_days_2023" > 150 AND counts."rainy_days_2023" < counts."rainy_days_2022" ORDER BY counts."rainy_days_2023" DESC NULLS LAST;
        \end{lstlisting} \\

        \midrule

        \multirow{18}{*}{\textbf{\modelName}} & \begin{lstlisting}[language=SQL]
WITH "rainy_2022" AS (  SELECT  g."stn" AS "usaf",  g."wban" AS "wban",  TRIM(s."name") AS "station_name",  COUNT(*) AS "rainy_days_2022"  FROM "NOAA_DATA"."NOAA_GSOD"."GSOD2022" g  JOIN "NOAA_DATA"."NOAA_GSOD"."STATIONS" s  ON g."stn" = s."usaf"  AND g."wban" = s."wban"  WHERE s."state" = 'WA'  AND g."stn" <> '999999'  AND g."prcp" IS NOT NULL  AND g."prcp" <> 99.99  AND g."prcp" > 0  GROUP BY g."stn", g."wban", TRIM(s."name") ), "rainy_2023" AS (  SELECT  g."stn" AS "usaf",  g."wban" AS "wban",  TRIM(s."name") AS "station_name",  COUNT(*) AS "rainy_days_2023"  FROM "NOAA_DATA"."NOAA_GSOD"."GSOD2023" g  JOIN "NOAA_DATA"."NOAA_GSOD"."STATIONS" s  ON g."stn" = s."usaf"  AND g."wban" = s."wban"  WHERE s."state" = 'WA'  AND g."stn" <> '999999'  AND g."prcp" IS NOT NULL  AND g."prcp" <> 99.99  AND g."prcp" > 0  GROUP BY g."stn", g."wban", TRIM(s."name") ) SELECT  r23."usaf" AS "station_id",  r23."station_name",  r23."rainy_days_2023",  r22."rainy_days_2022" FROM "rainy_2023" r23 JOIN "rainy_2022" r22 ON r23."usaf" = r22."usaf" AND r23."wban" = r22."wban" WHERE r23."rainy_days_2023" > 150 AND r23."rainy_days_2023" < r22."rainy_days_2022" ORDER BY r23."rainy_days_2023" DESC, r23."station_name";
        \end{lstlisting} \\

        \bottomrule
    \end{tabular}
    \caption{Qualitative Example 1 showcasing how \modelName{} improves over some of the previous approaches.}
    \label{tab:qual-examples-1}
\end{table*}

\begin{table*}[t]
    \centering
    \small
    \begin{tabular}{l|p{13.5cm}}
        \toprule
        \multicolumn{2}{c}{\textbf{Example 2}} \\
        \midrule
        \textbf{Question} & How many distinct pseudo users had positive engagement time in the 5-day period ending on January 7, 2021 at 23:59:59, but had no positive engagement time in the 2-day period ending on the same date (January 7, 2021 at 23:59:59) ? \\

        \midrule
        
        \multirow{17}{*}{\textbf{Spider-Agent}} & \begin{lstlisting}[language=SQL]
WITH engagement AS (  SELECT e."USER_PSEUDO_ID", \textbf{TO_DATE(e."EVENT_DATE", 'YYYYMMDD')} AS "evt_date"  FROM (  SELECT * FROM GA4.GA4_OBFUSCATED_SAMPLE_ECOMMERCE.EVENTS_20210103  UNION ALL SELECT * FROM GA4.GA4_OBFUSCATED_SAMPLE_ECOMMERCE.EVENTS_20210104  UNION ALL SELECT * FROM GA4.GA4_OBFUSCATED_SAMPLE_ECOMMERCE.EVENTS_20210105  UNION ALL SELECT * FROM GA4.GA4_OBFUSCATED_SAMPLE_ECOMMERCE.EVENTS_20210106  UNION ALL SELECT * FROM GA4.GA4_OBFUSCATED_SAMPLE_ECOMMERCE.EVENTS_20210107  ) e,  LATERAL FLATTEN (INPUT => e."EVENT_PARAMS") ep  WHERE e."EVENT_NAME" = 'user_engagement' AND ep.value:"key" = 'engagement_time_msec' AND COALESCE( ep.value:"value": "int_value"::INTEGER, 0) > 0), user_flags AS ( SELECT "USER_PSEUDO_ID", MAX(CASE WHEN "evt_date" BETWEEN '2021-01-03' AND '2021-01-07' THEN 1 ELSE 0 END) AS seven_day_positive, MAX(CASE WHEN "evt_date" BETWEEN '2021-01-06' AND '2021-01-07' THEN 1 ELSE 0 END) AS two_day_positive  FROM engagement GROUP BY "USER_PSEUDO_ID") SELECT COUNT(*) AS "distinct_pseudo_users" FROM user_flags WHERE seven_day_positive = 1 AND two_day_positive = 0;
        \end{lstlisting} \\

        \midrule

        \multirow{19}{*}{\textbf{ReFoRCE}} & \begin{lstlisting}[language=SQL]
WITH union_events AS (  SELECT * FROM  GA4.GA4_OBFUSCATED_SAMPLE_ECOMMERCE. "EVENTS_20210103"  UNION ALL  SELECT * FROM GA4.GA4_OBFUSCATED_SAMPLE_ECOMMERCE. "EVENTS_20210104"  UNION ALL  SELECT * FROM GA4.GA4_OBFUSCATED_SAMPLE_ECOMMERCE. "EVENTS_20210105"  UNION ALL  SELECT * FROM GA4.GA4_OBFUSCATED_SAMPLE_ECOMMERCE. "EVENTS_20210106"  UNION ALL  SELECT * FROM GA4.GA4_OBFUSCATED_SAMPLE_ECOMMERCE. "EVENTS_20210107" ), engagement AS (  SELECT  t."USER_PSEUDO_ID" AS user_id,  TO_DATE(t."EVENT_DATE", 'YYYYMMDD') AS event_date,  f.value:"value":"int_value"::INTEGER AS engagement_time_msec  FROM union_events t,  LATERAL FLATTEN(INPUT => t."EVENT_PARAMS") f  WHERE t."EVENT_NAME" = 'user_engagement'  AND f.value:"key"::STRING = 'engagement_time_msec'  AND f.value:"value":"int_value"::INTEGER > 0 ), users_5d AS (  SELECT DISTINCT user_id  FROM engagement  WHERE event_date BETWEEN '2021-01-03' AND '2021-01-07' ), users_2d AS (  SELECT DISTINCT user_id  FROM engagement  WHERE event_date BETWEEN '2021-01-06' AND '2021-01-07' ) SELECT COUNT(*) AS distinct_users FROM users_5d WHERE user_id NOT IN (SELECT user_id FROM users_2d);
        \end{lstlisting} \\

        \midrule
        
        \multirow{21}{*}{\textbf{\modelName}} & \begin{lstlisting}[language=SQL]
WITH "engaged_5d" AS (  SELECT DISTINCT "USER_PSEUDO_ID"  FROM (  SELECT "USER_PSEUDO_ID", "EVENT_PARAMS" FROM "GA4"."GA4_OBFUSCATED_SAMPLE_ECOMMERCE". "EVENTS_20210103" UNION ALL  SELECT "USER_PSEUDO_ID", "EVENT_PARAMS" FROM "GA4"."GA4_OBFUSCATED_SAMPLE_ECOMMERCE". "EVENTS_20210104" UNION ALL  SELECT "USER_PSEUDO_ID", "EVENT_PARAMS" FROM "GA4"."GA4_OBFUSCATED_SAMPLE_ECOMMERCE". "EVENTS_20210105" UNION ALL  SELECT "USER_PSEUDO_ID", "EVENT_PARAMS" FROM "GA4"."GA4_OBFUSCATED_SAMPLE_ECOMMERCE". "EVENTS_20210106" UNION ALL  SELECT "USER_PSEUDO_ID", "EVENT_PARAMS" FROM "GA4"."GA4_OBFUSCATED_SAMPLE_ECOMMERCE". "EVENTS_20210107"  ) t,  LATERAL FLATTEN(input => t."EVENT_PARAMS") ep  WHERE ep.value:key::string = 'engagement_time_msec'  AND COALESCE(ep.value:value:int_value,0)::number > 0 ), "engaged_2d" AS (  SELECT DISTINCT "USER_PSEUDO_ID"  FROM (  SELECT "USER_PSEUDO_ID", "EVENT_PARAMS" FROM "GA4"."GA4_OBFUSCATED_SAMPLE_ECOMMERCE"."EVENTS_20210106" UNION ALL  SELECT "USER_PSEUDO_ID", "EVENT_PARAMS" FROM "GA4"."GA4_OBFUSCATED_SAMPLE_ECOMMERCE"."EVENTS_20210107"  ) t,  LATERAL FLATTEN(input => t."EVENT_PARAMS") ep  WHERE ep.value:key::string = 'engagement_time_msec'  AND COALESCE(ep.value:value:int_value,0)::number > 0  ) SELECT COUNT(*) AS "distinct_pseudo_users" FROM "engaged_5d" WHERE "USER_PSEUDO_ID" NOT IN (SELECT "USER_PSEUDO_ID" FROM "engaged_2d");
        \end{lstlisting} \\
        \bottomrule
    \end{tabular}
    \caption{Qualitative Example 2 showcasing how \modelName{} improves over some of the previous approaches.}
    \label{tab:qual-examples-2}
\end{table*}

\subsection{Qualitative Examples}

To provide more insight where \modelName{} improves the existing models, we provide some qualitative samples in Table~\ref{tab:qual-examples-1} and Table~\ref{tab:qual-examples-2}.

\paragraph{Example 1}
Here, we note that Spider-Agent suffers from using the wrong table and completely misses the final results.
On the other hand, ReFoRCE and \modelName{} use the right table/schema; but ReFoRCE has additional entries and looser filters.
Additionally, Spider-Agent also uses station id to identify unique NOAA stations; while ReFoRCE and \modelName{} correctly use a combination USAF and WBAN for this identification.
Finally, the main logic issue is in the data filtering step. Spider-Agent only utilizes "precipitation" > 0 as the main filter.
ReFoRCE improves this by adding filters like "prcp" IS NOT NULL and "prcp" <> 99.99.
\modelName{} further improves over both by adding an additional filter of "stn" <> '999999' - which it procures after deep database exploration to be a placeholder station number.

\paragraph{Example 2}
First, in terms of query structure and readability, Spider-Agent uses nested subqueries and multiple CTEs, which makes it hard to follow.
ReFoRCE improves readability by modularizing unions, while \modelName{} is the most logically separated, with two explicit CTEs (engaged\_7d, engaged\_2d).
Creating these separate CTEs also helps to ensure better date filtering.
Spider-Agent uses implicit joins without clear separation of logic, while ReFoRCE and \modelName{} use LATERAL FLATTEN joins more explicitly, making semantics clearer.
Finally, the major distinction which leads to undercounting for Spider-Agent and ReFoRCE, relative to \modelName{} is the additional redundant filter/casting of date introduced in these SQL queries.
Instead, \modelName{} picks the date from the table name and avoids adding additional filters.

\subsection{Case Study of Parallelization}
\label{sec:case-study}

In Table~\ref{tab:case-study-deep}, we demonstrate an end-to-end case study for how PExA utilizes the multiple agents to explore one datapoint.
Overall, the trajectory has two planning/decomposition components from the planner.
In the first plan, two subplans are executed in parallel.
The first subplan is explored in 5-7 different manners and the final SQL is written for it.
For the second one, two execution strategies are explored and used.
In the second plan, one subplan is executed, for which four different explorations are conducted.
Finally, the proposer uses all this information to draft final SQL queries.
Multiple feedbacks from the SQL Executor help it to fix the query which finally is syntactically and semantically correct.
Finally, the planner reformats it a bit and returns as the final SQL.



\subsection{Updated SPIDER 2.0 Leaderboard}
\label{sec:leaderboard}

\begin{figure*}[h]
    \centering
    \includegraphics[width=0.9\linewidth]{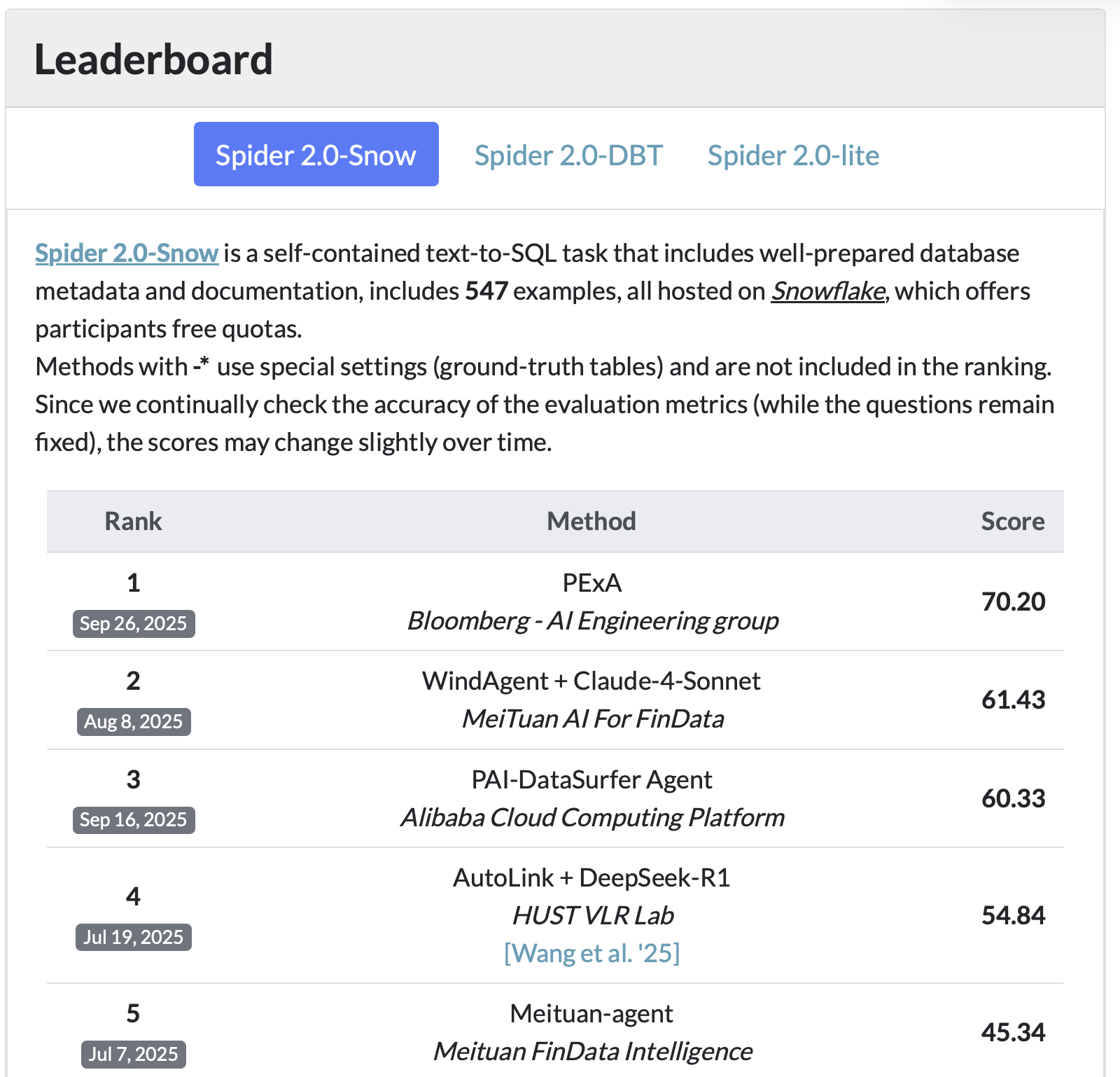}
    \caption{Snapshot of \modelName's performance compared to other methods as of November 15, 2025.}
    \label{fig:pexa-leaderboard}
\end{figure*}

In Figure~\ref{fig:pexa-leaderboard}, we provide a snapshot highlighting \modelName's state-of-the-art performance when it was updated on the main leaderboard for \dataName.
We also note how significantly \modelName{} outperforms all other models.

\section{Directions for Future Work}

In our work, we provide a strong proof-of-concept for parallelized exploration to improve the Pareto optimality of performance-latency in text-to-SQL on \dataName.
As part of future work, it would be imperative to provide the generalizability of this paradigm across other text-to-SQL datasets like SPIDER \cite{yu-etal-2018-spider} and BIRD \cite{bird}.
To this end, exploring post-training of individual components of our agentic workflow would be required.
Synthetic data generation methods \cite{duan-etal-2025-dsqg, parekh-etal-2025-snare} should also be explored to this end to  support this post-training.
At a broader scale, works can also explore utilizing this framework for broader set of tasks in code generation, tool-calling, and beyond \cite{wang2025swe, t2-bench, parekh-etal-2024-speed}.

\onecolumn
\begin{center}
\small
\begin{longtable}{l|p{13cm}}
\caption{Detailed Case Study of \modelName's exploration for a single datapoint.} \\
\toprule
\endfirsthead
\caption*{Table \thetable\ (continued)} \\
\toprule
\endhead
\bottomrule
\endfoot
\multirow{8}{*}{\textbf{Question}} & I want to know the GitHub project URLs for the top 3 most downloaded PyPI packages based on download count. First, extract PyPI package metadata including name, version, and project URLs. Filter these URLs to only include those that link to GitHub repositories. Use a regular expression to clean the GitHub URLs by removing unnecessary parts like 'issues', 'pull', 'blob', and 'tree' paths, keeping only the main repository URL. For packages with multiple versions, use only the most recent version based on upload time. Join this data with download metrics to determine the most downloaded packages. Return only the cleaned GitHub repository URLs (without quotation marks) for the top 3 packages by total download count, ensuring that only packages with valid GitHub URLs are included in the results. \\
        \midrule
        \multirow{3}{*}{\textbf{Planner Probes 1}} &['Retrieve a table listing PyPI package metadata with the columns: package name, version, project\_urls (or project\_url), and upload\_time.', 'Retrieve a table listing each PyPI package together with its total download count across all versions.'] \\
        \midrule
        \multirow{2}{*}{\textbf{Test Case 1}} & Retrieve a table listing PyPI package metadata with the columns: package name, version, project\_urls (or project\_url), and upload\_time. \\
        \midrule
        \multirow{26}{*}{\textbf{Execution 1}} & 1. SQL: SELECT "name" AS "package\_name", "version", "project\_urls", "upload\_time"
FROM "PYPI"."PYPI"."DISTRIBUTION\_METADATA"
WHERE "name" IS NOT NULL
LIMIT 100; \newline
Execution Output (truncated):
package\_name,version,project\_urls,upload\_time
sparganothis-vim,0.1.15,[],1720561327687084
sparganothis-vim,0.1.15,[],1720561419667601
safa,0.0.1,[],1720564065701497 \newline
2. SQL: WITH "downloads" AS (
 SELECT
 "project" AS "name",
 "file":"version"::string AS "version",
 MIN("timestamp") AS "first\_download\_ts"
 FROM "PYPI"."PYPI"."FILE\_DOWNLOADS"
 GROUP BY "project", "file":"version"::string
) SELECT
 d."name" AS "package\_name",
 d."version",
 d."project\_urls",
 d."upload\_time"
FROM "PYPI"."PYPI"."DISTRIBUTION\_METADATA" d
JOIN "downloads" f
 ON d."name" = f."name" AND d."version" = f."version"
LIMIT 100; \newline
Execution Output (truncated):
package\_name,version,project\_urls,upload\_time
segments-ai,1.0.1,[],1659525962095717
pyrish,0.3.1,[],1517356387930046
odoo9-addon-stock-quant-manual-assign,9.0.1.0.0.99.dev20,[],1500179644597289 \newline
3. SQL: SELECT
 d."name" AS "package\_name",
 d."version",
 d."project\_urls",
 d."upload\_time"
FROM "PYPI"."PYPI"."DISTRIBUTION\_METADATA" d
JOIN "PYPI"."INFORMATION\_SCHEMA"."TABLES" t
 ON t."TABLE\_NAME" = 'DISTRIBUTION\_METADATA'
LIMIT 100; \newline
Execution Output (truncated):
package\_name,version,project\_urls,upload\_time
sparganothis-vim,0.1.15,[],1720561327687084
sparganothis-vim,0.1.15,[],1720561419667601
safa,0.0.1,[],1720564065701497 \newline
4. SQL: SELECT
 "project" AS "package\_name",
 "file":"version"::string AS "version",
 NULL AS "project\_urls",
 MIN("timestamp") AS "upload\_time"
FROM "PYPI"."PYPI"."FILE\_DOWNLOADS"
GROUP BY "project", "file":"version"::string
LIMIT 100; \newline
Execution Output (truncated):
package\_name,version,project\_urls,upload\_time bs-solevaluator,0.1.1,,1649607460000000
django-graphql-auth,0.1.9,,1706310942000000
collabi,0.0.1,,1699984505000000 \\
        \midrule
        \multirow{5}{*}{\textbf{Execution 2}} & 1. SQL: SELECT "name" AS "package\_name", "version", "project\_urls" AS "project\_urls",
"upload\_time"
FROM "PYPI"."PYPI"."DISTRIBUTION\_METADATA"
LIMIT 10; \newline
Execution Output (truncated):
package\_name,version,project\_urls,upload\_time
sparganothis-vim,0.1.15,[],1720561327687084
sparganothis-vim,0.1.15,[],1720561419667601
safa,0.0.1,[],1720564065701497 \\
        \midrule
        \multirow{34}{*}{\textbf{Exploration 3}} & 1. SQL: SELECT COUNT(*) AS cnt\_non\_empty
FROM "PYPI"."PYPI"."DISTRIBUTION\_METADATA"
WHERE "project\_urls" IS NOT NULL AND ARRAY\_SIZE("project\_urls") > 0; \newline
Execution Output (truncated):
CNT\_NON\_EMPTY
42893 \newline
2. SQL: SELECT "name" AS "package\_name", "version", "project\_urls", "upload\_time"
FROM "PYPI"."PYPI"."DISTRIBUTION\_METADATA"
WHERE "project\_urls" IS NOT NULL AND ARRAY\_SIZE("project\_urls") > 0
LIMIT 10; \newline
Execution Output (truncated):
package\_name,version,project\_urls,upload\_time
renovosolutions.aws-cdk-aspects-security-group,1.0.214,"[ ""Source,
https://github.com/RenovoSolutions/cdk-aspects-library-security-group.git""
]",1661773668691461
connectome-interpreter,0.5.0,"[ ""Documentation,
https://connectome-interpreter.readthedocs.io/en/latest/"", ""Larva connectome
example,
https://colab.research.google.com/drive/1VIMNFBp7dCgN5XOQ9vvzPaqb80BGPZx4?usp=sharing"
", ""Adult connectome example (FAFB),
https://colab.research.google.com/drive/1ECUagwN-r2rnKyfcYgtR1oG8Lox8m8BW?usp=sharing"
" ]",1709899013955375
oomongo,1.1,"[ ""HomePage, https://github.com/lcctoor/arts/tree/main/arts/oomongo""
]",1709687704051481
selectolax,0.3.21,"[ ""Source code, https://github.com/rushter/selectolax""
]",1710085659536925
sweetrpg-model-core,0.0.154,"[ ""Documentation,
https://sweetrpg.github.io/model-core"" ]",1709610651959794
bolero,0.0.19,"[ ""Documentation, https://bolero.readthedocs.io/"", ""Source,
https://github.com/lhqing/bolero"", ""Home-page, https://github.com/lhqing/bolero""
]",1709615613500068
microagent,1.7.3,"[ ""Homepage, https://github.com/scailer/microagent"",
""Documentation, https://microagent.readthedocs.io/en/stable/"", ""Repository,
https://github.com/scailer/microagent.git"", ""Changelog,
https://github.com/scailer/microagent/blob/master/CHANGELOG.rst"" ]",1710257591115486
pysorters,0.5.6,"[ ""Source, https://github.com/yo1am1/pysorters""
]",1709687967377114
prot2d,0.4.1,"[ ""Repository, https://github.com/ConstantinCarl/prot2d"",
""Documentation, https://constantincarl.github.io/prot2d/"", ""Zenodo,
https://doi.org/10.5281/zenodo.10674045"", ""Example-Collab,
https://colab.research.google.com/drive/17u0twE81kYYspNFsdXUHrCyP33hj0dO6?usp=sharing"
" ]",1710169269545664
rcplus\_alloy\_common,3.7.0,"[ ""Repository,
https://github.com/ringier-data/rcplus-alloy-lib-py-common"" ]",1709301162301115 \newline
3. SQL: SELECT COUNT(*)
FROM "PYPI"."PYPI"."DISTRIBUTION\_METADATA"
WHERE "project\_url" IS NOT NULL; \newline
Failure Message:
SQL execution generated an error: \#\#ERROR\#\#000904 (42000): SQL compilation error:
error line 3 at position 6
invalid identifier '"project\_url"'
Please correct the query. \\
        \midrule
        \multirow{5}{*}{\textbf{Finish}} & \begin{lstlisting}[language=SQL]
SELECT
 "name" AS "package_name", "version", "project_urls", "upload_time"
FROM "PYPI"."PYPI"."DISTRIBUTION_METADATA"; 
        \end{lstlisting} \\
        \midrule
        \multirow{1}{*}{\textbf{Test Case 2}} &  Retrieve a table listing each PyPI package together with its total download count across all versions. \\
        \midrule
        \multirow{10}{*}{\textbf{Execution 1}} & 1. SQL: SELECT
 "project" AS "package\_name",
 COUNT(*) AS "total\_downloads"
FROM "PYPI"."PYPI"."FILE\_DOWNLOADS"
GROUP BY "project"
ORDER BY "total\_downloads" DESC; \newline
Execution Output (truncated):
package\_name,total\_downloads
s3transfer,2962234
attrs,1787211
colorama,1567249 \newline
2. SQL: SELECT
 dm."name" AS "package\_name",
 COUNT(*) AS "total\_downloads"
FROM "PYPI"."PYPI"."FILE\_DOWNLOADS" fd
JOIN (
 SELECT DISTINCT "name" FROM "PYPI"."PYPI"."DISTRIBUTION\_METADATA"
) dm
ON fd."project" = dm."name"
GROUP BY dm."name"
ORDER BY "total\_downloads" DESC; \newline
Execution Output (truncated):
package\_name,total\_downloads
wheel,1481586
tabulate,1478960
python-dateutil,1415163 \\
        \midrule
        \multirow{6}{*}{\textbf{Finish}} & \begin{lstlisting}[language=SQL]
SELECT
 "project" AS "package_name", COUNT(*) AS "total_downloads"
FROM "PYPI"."PYPI"."FILE_DOWNLOADS"
GROUP BY "project" ORDER BY "total_downloads" DESC; 
        \end{lstlisting} \\
        \midrule
        \multirow{2}{*}{\textbf{Planner Probes 2}} & ["Retrieve 10 sample rows from the PyPI distribution metadata where project\_urls contains 'github.com', returning package name, version, project\_urls, and upload\_time."] \\
        \midrule
        \multirow{2}{*}{\textbf{Test Case 3}} & Retrieve 10 sample rows from the PyPI distribution metadata where project\_urls contains 'github.com', returning package name, version, project\_urls, and upload\_time. \\
        \midrule
        \multirow{47}{*}{\textbf{Exploration 1}} & 1. SQL: SELECT "name", "version", "project\_urls", "upload\_time"
FROM "PYPI"."PYPI"."DISTRIBUTION\_METADATA"
WHERE "project\_urls"::string ILIKE '\%github.com\%'
LIMIT 10; \newline
Execution Output (truncated):
name,version,project\_urls,upload\_time
track-viz,0.5.0,"[ ""Changelog, https://github.com/JulienMBABD/track-viz/releases"",
""Documentation, https://track-viz.readthedocs.io"", ""Repository,
https://github.com/JulienMBABD/track-viz"" ]",1666370862718894
searchor,2.3.2,"[ ""Homepage, https://github.com/ArjunSharda/Searchor"", ""Bug
Tracker, https://github.com/ArjunSharda/Searchor/issues"" ]",1665751033856419
jpeglib,0.11.3,"[ ""Homepage, https://pypi.org/project/jpeglib/"",
""Documentation, https://jpeglib.readthedocs.io/en/latest/"", ""Source,
https://github.com/martinbenes1996/jpeglib/"" ]",1666259913352607 \newline
2. SQL: SELECT DISTINCT dm."name", dm."version", dm."project\_urls", dm."upload\_time"
FROM "PYPI"."PYPI"."DISTRIBUTION\_METADATA" dm
JOIN "PYPI"."PYPI"."FILE\_DOWNLOADS" fd
ON dm."name" = fd."project"
WHERE dm."project\_urls"::string ILIKE '\%github.com\%'
LIMIT 10; \newline
Execution Output (truncated):
name,version,project\_urls,upload\_time
iron-toolbox,1.0.45,"[ ""Bug Tracker,
https://github.com/IronTrainers/iron\_data\_toolbox/issues"" ]",1686764499727015
matplotlib,3.6.0rc2,"[ ""Documentation, https://matplotlib.org"", ""Source Code,
https://github.com/matplotlib/matplotlib"", ""Bug Tracker,
https://github.com/matplotlib/matplotlib/issues"", ""Forum,
https://discourse.matplotlib.org/"", ""Donate,
https://numfocus.org/donate-to-matplotlib"" ]",1661585954791952
cramjam,2.8.2,"[ ""homepage, https://github.com/milesgranger/pyrus-cramjam"",
""documentation, https://docs.rs/cramjam/latest/cramjam"", ""repository,
https://github.com/milesgranger/pyrus-cramjam"" ]",1709410989630295 \newline
3. SQL: SELECT dm."name", dm."version", dm."project\_urls", dm."upload\_time"
FROM "PYPI"."PYPI"."DISTRIBUTION\_METADATA" dm
JOIN "PYPI"."INFORMATION\_SCHEMA"."TABLES" t
ON t."TABLE\_NAME" = 'DISTRIBUTION\_METADATA' AND t."TABLE\_SCHEMA" = 'PYPI'
WHERE dm."project\_urls"::string ILIKE '\%github.com\%'
LIMIT 10; \newline
Execution Output (truncated):
name,version,project\_urls,upload\_time
BottleSessions,21.9.18,"[ ""repo, https://github.com/Glocktober/BottleSessions"",
""overview, https://github.com/Glocktober/BottleSessions/blob/main/README.md""
]",1632027763679760
PTX-now,2.1,"[ ""Bug Tracker, https://github.com/pypa/sampleproject/issues""
]",1631016956804108
aafragpy-serkol,0.8.45,"[ ""Bug Tracker,
https://github.com/skoldobskiy/aafragpy/issues"" ]",1632248257186784 \newline
4. SQL: SELECT DISTINCT dm."name", dm."version", dm."project\_urls", dm."upload\_time"
FROM "PYPI"."PYPI"."DISTRIBUTION\_METADATA" dm
JOIN "PYPI"."PYPI"."FILE\_DOWNLOADS" fd
ON dm."name" = fd."project"
JOIN "PYPI"."INFORMATION\_SCHEMA"."TABLES" t
ON t."TABLE\_NAME" = 'DISTRIBUTION\_METADATA' AND t."TABLE\_SCHEMA" = 'PYPI'
WHERE dm."project\_urls"::string ILIKE '\%github.com\%'
LIMIT 10; \newline
Execution Output (truncated):
name,version,project\_urls,upload\_time
magpylib,4.5.0rc0,"[ ""Bug Tracker, https://github.com/magpylib/magpylib/issues"",
""Changelog, https://github.com/magpylib/magpylib/blob/master/CHANGELOG.md"",
""Documentation, https://magpylib.readthedocs.io/en/latest/"", ""Repository,
https://github.com/magpylib/magpylib.git"" ]",1702456277638981
duckdb,0.5.2.dev1120,"[ ""Documentation,
https://duckdb.org/docs/api/python/overview"", ""Source,
https://github.com/duckdb/duckdb/blob/master/tools/pythonpkg"", ""Issues,
https://github.com/duckdb/duckdb/issues"", ""Changelog,
https://github.com/duckdb/duckdb/releases"" ]",1666636910274206
augraphy,8.2.3,"[ ""Bug Tracker, https://github.com/sparkfish/augraphy/issues""
]",1682936295648019 \\
        \midrule
        \multirow{6}{*}{\textbf{Finish}} & \begin{lstlisting}[language=SQL]
SELECT "name", "version", "project_urls", "upload_time"
FROM "PYPI"."PYPI"."DISTRIBUTION_METADATA"
WHERE "project_urls"::string ILIKE '%github.com%'
LIMIT 10; 
        \end{lstlisting} \\
        \midrule
        \textbf{Planner} & Call Proposer \\
        \midrule
        \multirow{52}{*}{\textbf{Proposer 1}} & \begin{lstlisting}[language=SQL]
-- Assumptions & reasoning in comments:-- 1."PYPI.PYPI.DISTRIBUTION_METADATA" keeps one row per (package, version) and holds anARRAY variant column "project_urls" -- whose elements are strings like"Source, https://github.com/org/repo" (possibly ending in .git or deeper paths).--2. We only want the *latest* version per package (highest "upload_time").-- 3. Weonly keep packages whose latest metadata contains at least one GitHub URL.-- 4.Using that latest-metadata list, we flatten the array, extract the first GitHub URLcontained in every element and-- clean it so that only the canonical repo rootremains (https://github.com/<owner>/<repo>), removing:-- trailing pathfragments such as /issues, /pull/123, /blob/main/README.md, /tree/master, etc.-- a trailing ".git"-- a trailing slash-- 5. We join this per-packagecleaned URL to the download totals derived from "PYPI.PYPI.FILE_DOWNLOADS".-- 6.Finally, we return the cleaned GitHub URLs (no quotes around them in SQL output) forthe top 3 packages by total download count.WITH "downloads" AS ( -- totaldownloads for every package (all versions) SELECT "project"AS "package_name", COUNT(*) AS"total_downloads" FROM "PYPI"."PYPI"."FILE_DOWNLOADS" GROUP BY"project"),-- latest metadata row per package (most recentupload_time)"latest_metadata" AS ( SELECT "name" AS"package_name", "project_urls", ROW_NUMBER() OVER (PARTITION BY"name" ORDER BY "upload_time" DESC) AS "rn" FROM"PYPI"."PYPI"."DISTRIBUTION_METADATA" WHERE "project_urls" IS NOT NULLAND "project_urls"::string ILIKE '\%github.com\%'),"latest_with_urls" AS (SELECT lm."package_name", lm."project_urls" FROM "latest_metadata" lmWHERE lm."rn" = 1 -- keep only the most-recent version perpackage),-- flatten project_urls array and pull out raw GitHubURLs"flattened" AS ( SELECT lwu."package_name",TRIM(f.value::string) AS "url_element" FROM "latest_with_urls" lwu,LATERAL FLATTEN(input => lwu."project_urls") f),"raw_github_urls" AS (SELECT "package_name", -- extract the *first* substring that lookslike a GitHub URL REGEXP_SUBSTR("url_element",\texttt{'https?://github\\.com/[^\\s,"\\)]+')} AS "raw_url" FROM "flattened"WHERE REGEXP_LIKE("url_element", 'https?://github\\.com/')),-- clean theGitHub URLs down to repo root"cleaned_urls" AS ( SELECT DISTINCTrgu."package_name", -- 1) keep only scheme+owner/repo, discardingeverything afterwards -- 2) drop trailing .git if present -- 3) droptrailing slash REGEXP_REPLACE( REGEXP_REPLACE(LOWER(rgu."raw_url"), -- normalise case\texttt{'^https?://github\\.com/([^/]+/[^/]+)/?.*$', -- capture owner/repoonly 'https://github.com/\\\\1' )},'\\\\.git/?$', -- remove trailing.git (with optional /) '' ) AS "cleaned_url" FROM"raw_github_urls" rgu WHERE rgu."raw_url" IS NOT NULL),-- attachdownload counts and pick the main URL per package (if multiple URLs, picklexicographically smallest)"package_with_downloads" AS ( SELECTd."package_name", d."total_downloads", MIN(cu."cleaned_url")AS "cleaned_url" -- deterministic pick per package FROM "downloads" dJOIN "cleaned_urls" cu ON d."package_name" = cu."package_name" GROUPBY d."package_name", d."total_downloads"),-- top-3 by downloadcount"top3" AS ( SELECT * FROM "package_with_downloads" ORDER BY"total_downloads" DESC LIMIT 3)SELECT "cleaned_url"FROM"top3"ORDER BY "total_downloads" DESC;
\end{lstlisting} \\
        \midrule
        \multirow{2}{*}{\textbf{SQL Executor}} & 
        SQL executed, but no valid results could be extracted. There seems to be some logical error in the SQL query. Please correct the query. \\
        \multirow{44}{*}{\textbf{Proposer 2}} & \begin{lstlisting}[language=SQL]
-- Goal: GitHub repository URLs for the 3 most-downloaded PyPI packages(by total file downloads)-- Steps--------------------------------------------------------------- 1. Get themost-recent metadata row (highest upload_time) per package.-- 2. FLATTEN theproject_urls (array / object) and keep only those elements that mentiongithub.com.-- 3. Extract the raw GitHub URL substring from every element, thennormalise it to the canonical-- repo root "https://github.com/<owner>/<repo>"(remove /issues, /blob/..., /tree/..., *.git, etc.).-- 4. Aggregate to one(deterministic) cleaned URL per package.-- 5. Join with total download counts andreturn the top-3 packages that have a valid GitHubURL.--------------------------------------------------------------------- WITH latest_metadata AS ( SELECT "name" AS "package_name","project_urls", ROW_NUMBER() OVER (PARTITION BY "name" ORDER BY"upload_time" DESC) AS "rn" FROM"PYPI"."PYPI"."DISTRIBUTION_METADATA" WHERE "project_urls" IS NOT NULL-- only rows that actually contain project_urls),latest AS ( SELECT"package_name", "project_urls" FROM latest_metadata WHERE "rn" = 1-- keep only the newest version per package),flattened AS ( -- explode thearray/object so each URL (or key-value) becomes its own row SELECTl."package_name", f.value::string AS "url_element" FROM latestl, LATERAL FLATTEN(input => l."project_urls") f WHERE f.value ISNOT NULL AND f.value::string ILIKE '%github.com%'),github_raw AS (-- pull out the first GitHub URL substring from each element SELECT DISTINCT"package_name", REGEXP_SUBSTR( "url_element",'https?://github\\.com/[^\\s)"\\'']+' -- stops at space, ), " or ' )AS "raw_url" FROM flattened WHERE REGEXP_LIKE("url_element",'https?://github\\.com/')),cleaned_urls AS ( -- 1) drop trailing .git(optionally followed by /) -- 2) collapse to repo root by capturing justowner/repo and rebuilding the URL SELECT "package_name",REGEXP_REPLACE( REGEXP_REPLACE(LOWER("raw_url"), '\\\\.?git/?$', ''),-- step 1 '^https?://github\\.com/([^/]+/[^/]+)(?:/.*)?$', --step 2 (capture) 'https://github.com/\\\\1' ) AS"cleaned_url" FROM github_raw WHERE "raw_url" IS NOTNULL),per_package_url AS ( -- choose a single deterministic URL per package(lexicographically smallest) SELECT "package_name",MIN("cleaned_url") AS "cleaned_url" FROM cleaned_urls GROUP BY"package_name"),downloads AS ( -- total file downloads (all versions) perpackage SELECT "project" AS "package_name", COUNT(*) AS"total_downloads" FROM "PYPI"."PYPI"."FILE_DOWNLOADS" GROUP BY"project"),ranked AS ( SELECT d."total_downloads",u."cleaned_url" FROM downloads d JOIN per_package_url u ONd."package_name" = u."package_name" ORDER BY d."total_downloads" DESCLIMIT 3)SELECT "cleaned_url"FROM rankedORDER BY "total_downloads"DESC;
        \end{lstlisting} \\
        \midrule
        \multirow{3}{*}{\textbf{SQL Executor}} & SQL execution generated an error \#\#ERROR\#\#001003 (42000): SQL compilation error: syntax error line 33 at position 43 unexpected ']'. syntax error line 34 at position 10 unexpected 'AS'. Please correct the query. \\
        \midrule
        \multirow{40}{*}{\textbf{Proposer 3}} & \begin{lstlisting}[language=SQL]
-- GitHub repository URLs for the 3 most-downloaded PyPI packages ----------------------------------------------------------------------- 1. Identifythe most-recent metadata row for every package (by upload_time).-- 2. FLATTEN"project_urls" and keep only the elements that reference GitHub.-- 3. Extract theraw GitHub URL substring, then normalise it to the repo root--(https://github.com/<owner>/<repo>), removing trailing path fragments-- and apossible ".git".-- 4. Join these packages to their total download counts and pickthe top 3.------------------------------------------------------------------------ WITH "latest_metadata" AS ( SELECT "name" AS"package_name", "project_urls", ROW_NUMBER() OVER (PARTITION BY"name" ORDER BY "upload_time" DESC) AS "rn" FROM"PYPI"."PYPI"."DISTRIBUTION_METADATA" WHERE "project_urls" IS NOTNULL),"latest" AS ( SELECT "package_name", "project_urls" FROM"latest_metadata" WHERE "rn" = 1 -- keep only thenewest version per package),"flattened" AS ( -- explode the project_urlsarray/object SELECT l."package_name", f.value::string AS"url_element" FROM "latest" l, LATERAL FLATTEN(input =>l."project_urls") f WHERE f.value IS NOT NULL AND f.value::stringILIKE '%github.com%'),"github_raw" AS ( -- extract the GitHub URLsubstring (up to first whitespace) SELECT DISTINCT "package_name",REGEXP_SUBSTR( "url_element",'https?://github\\.com/[^\\\\s]+' ) AS "raw_url" FROM "flattened"WHERE REGEXP_LIKE("url_element", 'https?://github\\.com/')),"cleaned_urls"AS ( -- normalise to canonical repo root and strip trailing .git SELECT"package_name", REGEXP_REPLACE(REGEXP_REPLACE(LOWER("raw_url"), '\\\\.git/?$', ''), -- drop .git'^https?://github\\.com/([^/]+/[^/]+)(/.*)?$', -- keep owner/repoonly 'https://github.com/\\\\1' ) AS "cleaned_url" FROM"github_raw" WHERE "raw_url" IS NOT NULL),"per_package_url" AS (-- deterministic single URL per package (lexicographically smallest) SELECT"package_name", MIN("cleaned_url") AS "cleaned_url" FROM"cleaned_urls" GROUP BY "package_name"),"downloads" AS ( --total file downloads per package (all versions) SELECT "project" AS"package_name", COUNT(*) AS "total_downloads" FROM"PYPI"."PYPI"."FILE_DOWNLOADS" GROUP BY "project"),"ranked" AS( SELECT d."total_downloads", u."cleaned_url" FROM"downloads" d JOIN "per_package_url" u ON d."package_name" =u."package_name" ORDER BY d."total_downloads" DESC LIMIT 3)SELECT"cleaned_url"FROM "ranked"ORDER BY "total_downloads" DESC;
        \end{lstlisting} \\
        \midrule
        \multirow{2}{*}{\textbf{SQL Executor}} & SQL executed, but no valid results could be extracted. There seems to be some logical error in the SQL query. Please correct the query. \\
        \midrule
        \multirow{40}{*}{\textbf{Proposer 4}} & \begin{lstlisting}[language=SQL]
-- GitHub repository URLs for the three most-downloaded PyPIpackages -- --------------------------------------------------------------------- 1. Find the latest (by upload_time) metadata row for every package.-- 2. From thoserows, FLATTEN the "project_urls" array / object and keep-- only elementsmentioning GitHub.-- 3. Extract the repository-root URL (https://github.com/<owner>/<repo> ),-- stripping any trailing ".git","/issues", "/pull/...", etc.-- 4. Keep one deterministic URL per package, joinwith total download-- counts, and return the URLs of the top-3packages. ------------------------------------------------------------------------ WITH latest_metadata AS ( SELECT "name" AS"package_name", "project_urls", ROW_NUMBER() OVER (PARTITION BY"name" ORDER BY "upload_time" DESC) AS "rn" FROM"PYPI"."PYPI"."DISTRIBUTION_METADATA" WHERE "project_urls" IS NOT NULL),latest AS ( SELECT "package_name", "project_urls" FROM latest_metadata WHERE "rn" = 1 -- newest version perpackage),flattened AS ( SELECT l."package_name",f.value::string AS "url_element" FROM latest l, LATERALFLATTEN(input => l."project_urls") f WHERE f.value IS NOT NULL ANDf.value::string ILIKE '%github.com%'),cleaned_urls AS ( /* 1)Extract the substring up to repo root (owner/repo). 2) Remove a trailing".git" (with optional slash). Examples handled: -https://github.com/tensorflow/tensorflow.git -> https://github.com/tensorflow/tensorflow -https://github.com/pallets/flask/tree/main -> https://github.com/pallets/flask - Source,https://github.com/psf/requests/issues -> https://github.com/psf/requests*/ SELECT DISTINCT "package_name", /* step-A: repo rootsubstring */ REGEXP_SUBSTR(LOWER("url_element"),'https?://github\\\\.com/[^/]+/[^/]+') AS "root_url" FROMflattened),normalised_urls AS ( SELECT "package_name",-- step-B: drop trailing .git and any trailing slashRTRIM(REGEXP_REPLACE("root_url", '\\\\.git$', ''), '/') AS "cleaned_url" FROMcleaned_urls WHERE "root_url" IS NOT NULL),per_package_url AS ( --choose a single deterministic URL per package (lexicographically smallest) SELECT"package_name", MIN("cleaned_url") AS "cleaned_url" FROMnormalised_urls GROUP BY "package_name"),downloads AS ( SELECT"project" AS "package_name", COUNT(*) AS "total_downloads" FROM"PYPI"."PYPI"."FILE_DOWNLOADS" GROUP BY "project"),ranked AS (SELECT d."total_downloads", u."cleaned_url" FROM downloads d JOINper_package_url u ON d."package_name" = u."package_name" ORDER BYd."total_downloads" DESC LIMIT 3)SELECT "cleaned_url"FROMrankedORDER BY "total_downloads" DESC;
        \end{lstlisting} \\
        \midrule
        \textbf{SQL Executor} & SQL query executed successfully! \\
        \midrule
        \textbf{Semantic Verifier} & The SQL query is semantically correct! \\
        \midrule
        \multirow{60}{*}{\textbf{Planner Finish}} & \begin{lstlisting}[language=SQL]
-- GitHub repository URLs for the three most-downloaded PyPI packages
-- -------------------------------------------------------------------
-- 1. Find the latest (by upload_time) metadata row for every package.
-- 2. From those rows, FLATTEN the "project_urls" array / object and keep
-- only elements mentioning GitHub.
-- 3. Extract the repository-root URL ( https://github.com/<owner>/<repo> ),
-- stripping any trailing ".git", "/issues", "/pull/...", etc.
-- 4. Keep one deterministic URL per package, join with total download
-- counts, and return the URLs of the top-3 packages.
------------------------------------------------------------------------
WITH latest_metadata AS (
 SELECT "name" AS "package_name", "project_urls", ROW_NUMBER() OVER (PARTITION BY "name" ORDER BY "upload_time" DESC) AS "rn"
 FROM "PYPI"."PYPI"."DISTRIBUTION_METADATA"
 WHERE "project_urls" IS NOT NULL
),
latest AS (
 SELECT "package_name", "project_urls"
 FROM latest_metadata
 WHERE "rn" = 1 -- newest version per package
),
flattened AS (
 SELECT l."package_name", f.value::string AS "url_element"
 FROM latest l, LATERAL FLATTEN(input => l."project_urls") f
 WHERE f.value IS NOT NULL AND f.value::string ILIKE '%github.com%'
),
cleaned_urls AS (
 /*
 1) Extract the substring up to repo root (owner/repo).
 2) Remove a trailing ".git" (with optional slash).
 */
 SELECT DISTINCT "package_name", /* step-A: repo root substring */ REGEXP_SUBSTR(LOWER("url_element"), 'https?://github\\.com/[^/]+/[^/]+') AS "root_url"
 FROM flattened
),
normalised_urls AS (
 SELECT "package_name", -- step-B: drop trailing .git and any trailing slash RTRIM(REGEXP_REPLACE("root_url", '\\.git$', ''), '/') AS "cleaned_url"
 FROM cleaned_urls
 WHERE "root_url" IS NOT NULL
),
per_package_url AS (
 SELECT "package_name", MIN("cleaned_url") AS "cleaned_url"
 FROM normalised_urls
 GROUP BY "package_name"
),
downloads AS (
 SELECT "project" AS "package_name", COUNT(*) AS "total_downloads"
 FROM "PYPI"."PYPI"."FILE_DOWNLOADS"
 GROUP BY "project"
),
ranked AS (
 SELECT d."total_downloads", u."cleaned_url"
 FROM downloads d
 JOIN per_package_url u ON d."package_name" = u."package_name"
 ORDER BY d."total_downloads" DESC LIMIT 3
)

SELECT "cleaned_url"
FROM ranked
ORDER BY "total_downloads" DESC;
        \end{lstlisting} \\
\label{tab:case-study-deep}
\end{longtable}
\end{center}
\twocolumn

\end{document}